\pdfoutput=1

\documentclass[sigplan, 10pt]{acmart}
\AtBeginDocument{%
  \providecommand\BibTeX{{%
    \normalfont B\kern-0.5em{\scshape i\kern-0.25em b}\kern-0.8em\TeX}}}

\copyrightyear{2023}
\acmYear{2023}
\setcopyright{acmcopyright}
\acmDOI{10.1145/3552326.3567501}

\acmConference[EuroSys '23]{Eighteenth European Conference on Computer Systems}{May 8--12, 2023}{Rome, Italy}
\acmBooktitle{Eighteenth European Conference on Computer Systems
	(EuroSys '23), May 8--12, 2023, Rome,
	Italy}
\acmPrice{15.00}
\acmISBN{978-1-4503-9487-1/23/05}

\usepackage{balance}
\usepackage{enumitem}
\usepackage{multirow}
\usepackage{subcaption}
\usepackage[linesnumbered, noend, ruled, vlined]{algorithm2e}
\usepackage{xspace}
\usepackage[a-1b]{pdfx}
\usepackage[compact]{titlesec}

\newcommand{\datastructure}{DENSE\xspace}
\newcommand{\systemname}{MariusGNN\xspace}
\newcommand{\replacementpolicy}{COMET\xspace}

\newcommand{\newparagraph}[1]{{\vspace{2pt}\noindent\textbf{#1}}}

\begin{document}
\title{\systemname: Resource-Efficient Out-of-Core Training of Graph Neural Networks}


\author{Roger Waleffe}
\authornote{Corresponding author: waleffe@wisc.edu}
\affiliation{%
	\institution{University of Wisconsin--Madison}
	\city{}
	\state{}
	\country{}
}
\author{Jason Mohoney}
\affiliation{%
	\institution{Universtity of Wisconsin--Madison}
	\city{}
	\state{}
	\country{}
}
\author{Theodoros Rekatsinas}
\affiliation{%
	\institution{ETH Z{\"u}rich}
	\city{}
	\state{}
	\country{}
}
\author{Shivaram Venkataraman}
\affiliation{%
	\institution{University of Wisconsin--Madison}
	\city{}
	\state{}
	\country{}
}

\renewcommand{\shortauthors}{Waleffe, et al.}

\begin{abstract}
We study training of Graph Neural Networks (GNNs) for large-scale graphs. We revisit the premise of using distributed training for billion-scale graphs and show that for graphs that fit in main memory or the SSD of a single machine, out-of-core pipelined training with a single GPU can outperform state-of-the-art (SoTA) multi-GPU solutions. We introduce \systemname, the first system that utilizes the entire storage hierarchy---including disk---for GNN training. \systemname introduces a series of data organization and algorithmic contributions that 1) minimize the end-to-end time required for training and 2) ensure that models learned with disk-based training exhibit accuracy similar to those fully trained in memory. We evaluate \systemname against SoTA systems for learning GNN models and find that single-GPU training in \systemname achieves the same level of accuracy up to 8$\times$ faster than multi-GPU training in these systems, thus, introducing an order of magnitude monetary cost reduction. \systemname is open-sourced at \url{www.marius-project.org}.
\end{abstract}




\maketitle

\section{Introduction}
Graph Neural Networks (GNNs) have emerged as an important method for machine learning (ML)~\cite{chami2021machine}. A diverse array of models~\cite{graphsage, gat, distmult, transe} have been introduced to apply ML over protein structures~\cite{STROKACH2020402}, social networks~\cite{fan2019graph}, and knowledge graphs~\cite{park2019estimating}. These models achieve state-of-the-art (SoTA) accuracy for \emph{node classification} and \emph{link prediction}~\cite{kipf2016semi, shang2021discrete, zhang2018link} by encoding data dependencies that arise in graphs.

\begin{table}[t]\footnotesize
    \caption{Many of the large graphs that GNNs are applied to can fit in main memory or the disk of a single commodity machine (e.g., AWS P3 GPU instances range from 61-488GB of CPU memory and contain up to 16TB of disk storage). The first five graphs are publicly available.}
    \vspace{-0.1in}
    \label{tab:intro_table}
    \begin{tabular}{l c c c c c c}
        \toprule
        \multirow{2}{*}[-.5ex]{Graph} & 
        \multirow{2}{*}[-.5ex]{Nodes} & 
        \multirow{2}{*}[-.5ex]{Edges} &
        \multirow{2}{1.8em}[-.5ex]{Feat. Dim} & 
        \multicolumn{3}{c}{Memory (GB)}\\
        \cmidrule(lr){5-7}
        &&&& Edges & Feat. & Tot.\\ 
        \midrule
        Papers100M~\cite{hu2020ogb} & 111M & 1.62B & 128 & 13 & 57 & 70\\
        Mag240M-Cites~\cite{hu2021ogblsc} & 122M & 1.30B & 768 & 10 & 375 & 385\\
        Freebase86M~\cite{freebase} & 86M & 338M & 100 & 4 & 69 & 73\\
        WikiKG90Mv2~\cite{hu2021ogblsc} & 91M & 601M & 100 & 7 & 73 & 80\\
        Hyperlink 2012~\cite{hyperlink} & 3.5B & 128B & 50 & 2k & 1.4k & 3.4k\\
        Facebook15~\cite{ching2015one} & 1.4B & 1T & 100 & 8k & 560 & 8.5k\\
        \bottomrule
    \end{tabular}
    \vspace{-0.1in}
\end{table}

However, resource-efficient GNN training over large-scale graphs has been highlighted as a major challenge in the literature~\cite{chami2021machine, zheng2022distributed, dorylus, p3gnn}. Graphs used in production settings contain millions of nodes and billions of edges. In addition, their nodes and edges are associated with feature vectors that form the inputs to GNNs. This data can require hundreds of GBs of storage~\cite{ilyas2022saga, mosaic} (Table~\ref{tab:intro_table}), which exceeds the capacity of GPU accelerators (e.g., 16GB on NVIDIA V100s). Moreover, the node representations at internal GNN layers depend on the nodes' multi-hop neighborhood, thus, a GNN dataflow graph can scale exponentially as more layers are added in a GNN. These two characteristics---graph storage sizes and GNN neighborhood dependencies---necessitate the use of mini-batch training coupled with multi-hop neighbor sampling for learning GNNs over large-scale inputs~\cite{dong2021global, largegcn}.

To perform training over large graphs, SoTA systems, such as Deep Graph Library (DGL)~\cite{dgl} and PyTorch Geometric (PyG)~\cite{pyg}, rely on CPU memory for graph storage and use distributed mini-batch training either over multiple GPUs or multiple machines. However, distributed solutions introduce deployment and maintenance overheads~\cite{mcsherry2015scalability} and in many cases obtain sub-linear speedups~\cite{zheng2022distributed}. For example, we find that DGL yields only a 2.2$\times$ speedup when using eight instead of one GPU (see Section~\ref{subsec:e2e_comp}). As a result, current distributed systems can incur increased monetary costs due to underutilization of allocated hardware.

We revisit GNN training and ask: \emph{When is distributed GNN training needed?} We argue that for graphs that fit in memory or the SSD of a single machine (Table~\ref{tab:intro_table}), \emph{out-of-core pipelined training on a single machine can outperform SoTA distributed systems}. We introduce \systemname, a system that utilizes the entire memory hierarchy---including disk---for GNN training over large-scale graphs. While prior work including GraphChi~\cite{graphchi}, Mosaic~\cite{mosaic}, and COST~\cite{mcsherry2015scalability}, have proposed using a single machine with disk for processing large graphs, they focus on graph analytics workloads such as PageRank. More recent graph ML systems such as Marius~\cite{mohoney2021marius} and PyTorch BigGraph~\cite{pytorchbiggraph} utilize disk, but do not contain multi-hop neighborhood samplers that are necessary for GNNs and thus only support specialized link prediction models for knowledge graph embeddings. In contrast, \systemname supports disk-based training of general $k$-hop sampling-based GNN models (subsuming prior graph ML systems) for both node classification and link prediction.

With experiments over four datasets using popular GNN architectures, we show that \systemname's disk-based, single-GPU training can be 8$\times$ faster than eight-GPU deployments of SoTA systems. This improvement yields monetary cost reductions of an order of magnitude. We find that for graphs where SoTA systems can take six days and \$1720 dollars to train a GNN, \systemname needs only eight hours and \$36 dollars for training, a $48\times$ reduction in monetary cost (WikiKG90Mv2, Table~\ref{tab:sys_comparison_lp}). Moreover, we show that single-GPU training can be sufficient for large-scale graphs: \textit{We use MariusGNN to train a GNN over the entire hyperlink graph from the Common Crawl 2012 web corpus, a graph with 3.5B nodes (web pages) and 128B edges (hyperlinks between pages) (Table~\ref{tab:intro_table})}. MariusGNN can learn vector representations for all \textit{3.5B nodes using only a single machine with one GPU, 60GB of RAM, and a large SSD, leading to a cost of just \$564/epoch}. Achieving these results required innovation and careful engineering along two dimensions: 1) optimized sampling to minimize the exponential costs of mini batch construction and 2) novel policies for transferring and processing training data that can maximize disk throughput while ensuring model accuracy is comparable to full in-memory training.

Despite recent efforts for efficient sampling when constructing a mini batch~\cite{chen2018fastgcn, zou2019layer, lazygcn, zeng2020graphsaint, clustergcn}, we find that existing sampling algorithms in SoTA systems bottleneck pipelined training and lead to the underutilization of GPUs. For example, using PyG's CPU-based neighborhood sampler for a three-layer GNN on a graph with 100M nodes takes 1200ms while the GPU-based operations for training take only 170ms (Section~\ref{subsec:dense_eval}). These timings arise because current approaches for sampling multi-hop neighborhoods resample one-hop neighbors for graph nodes multiple times while constructing a single GNN dataflow graph (e.g., Figure~\ref{fig:background_fig}). Due to the graph structure, the same nodes can appear repeatedly both within a layer and across different layers of the dataflow graph. When a node appears multiple times in the same layer, existing systems (e.g., DGL~\cite{dgl}) only sample the one-hop neighbors of this node once, yet when a node appears in separate layers, the one-hop neighbors of this node are resampled each time (e.g., node A in Figure~\ref{fig:background_fig}). Such resampling leads to redundant computation and data transfers and can limit pipeline throughput when training multi-layer GNNs.

In \systemname, we introduce a new data structure to minimize redundant computation and data transfers during sampling (Section~\ref{sec:mini_batch}). We term the new data structure \emph{\datastructure}, as it uses a Delta Encoding of Neighborhood SamplEs for nodes in successive GNN layers. \datastructure allows us to reuse sampled one-hop neighbors to construct the input of different GNN layers and enables in-CPU sampling that is up to 14$\times$ faster than SoTA systems (Section~\ref{subsec:dense_eval}). \datastructure also enables GNN forward pass computations up to 8$\times$ faster than competing systems by utilizing optimized dense GPU kernels rather than custom kernels for sparse matrix representations.

Disk-based execution in \systemname requires that the graph be split into partitions and subsets of these partitions be transferred to CPU memory for mini-batch training. We study partition replacement policies and mini batch construction policies for both link prediction and node classification.

For link prediction, we show that prior SoTA partition replacement policies, which greedily minimize IO, lead to biased models and harm accuracy when training GNNs. For example, when we implement the Buffer-aware Edge Traversal Algorithm (BETA) from prior work~\cite{mohoney2021marius} in \systemname, we find that the accuracy of learned GNN models drops by up to 16\% on common benchmarks when compared to training with the entire graph in main memory. We study the reason behind this drop (Section~\ref{subsec:disk_lp}) and find that existing policies lead to consecutive mini batches containing correlated training examples, a property that conflicts with the independently distributed assumption of ML training data.

To close the accuracy gap between in-memory and out-of-core training while minimizing disk IO, we introduce a new policy termed \replacementpolicy (COrrelation Minimizing Edge Traversal). First, \replacementpolicy separates the granularity of data storage and access from data transfer by utilizing two levels of partitioning (physical and logical partitions). Second, it decouples mini batch generation from partition replacement, allowing for randomized deferred processing of training examples, thus, further shuffling the order in which training examples are processed. To maximize throughput and accuracy, we provide automated rules for setting \replacementpolicy's hyperparameters. Using \replacementpolicy, we are able to reduce the gap between the link prediction accuracy of disk-based training and that of training with the full graph in memory by up to 80\% compared to prior work (Section~\ref{sec:eval_disk_pol}).

For node classification, we find that simple data transfer policies can be effective. In large-scale graphs ~\cite{hu2021ogblsc, hu2020ogb}, it is often the case that the set of nodes labeled for training is only 1-10\% of all graph vertices. Based on this information, we show that a policy that caches all labeled nodes in main memory and transfers random partitions from disk at the beginning of every epoch yields high-accuracy models for all the benchmark graphs we consider (Section~\ref{sec:eval_disk_pol}).

\section{Preliminaries on Training GNNs}
\label{sec:prelim}

We focus on training GNNs for node classification and link prediction. Given a graph $G= (V,E)$ with nodes $V$ and edges $E$, node classification assigns a class label to a node $v \in V$ given its local information (e.g., local features of the node) and information from its neighborhood in $G$. Link prediction considers a pair of nodes $(v_1, v_2) \in V \times V$ and predicts if the two nodes should be connected via an edge. Link prediction also uses the local features of nodes $v_1$ and $v_2$ and information from their neighborhoods to obtain a prediction.

Local information for each node in $G$ is modeled with a \emph{base vector representation} that encodes local node features. This representation can either be fixed (e.g., the conference of a paper in a citation graph) or \emph{learned} (e.g., a learned representation of the type of an entity node in a knowledge graph). For a node $v \in V$, we denote this base vector representation $h_v^0$. Base representations are stored in a \emph{lookup table}. 

\begin{figure}[!t]
  \centering
  \includegraphics[width=0.41\textwidth]{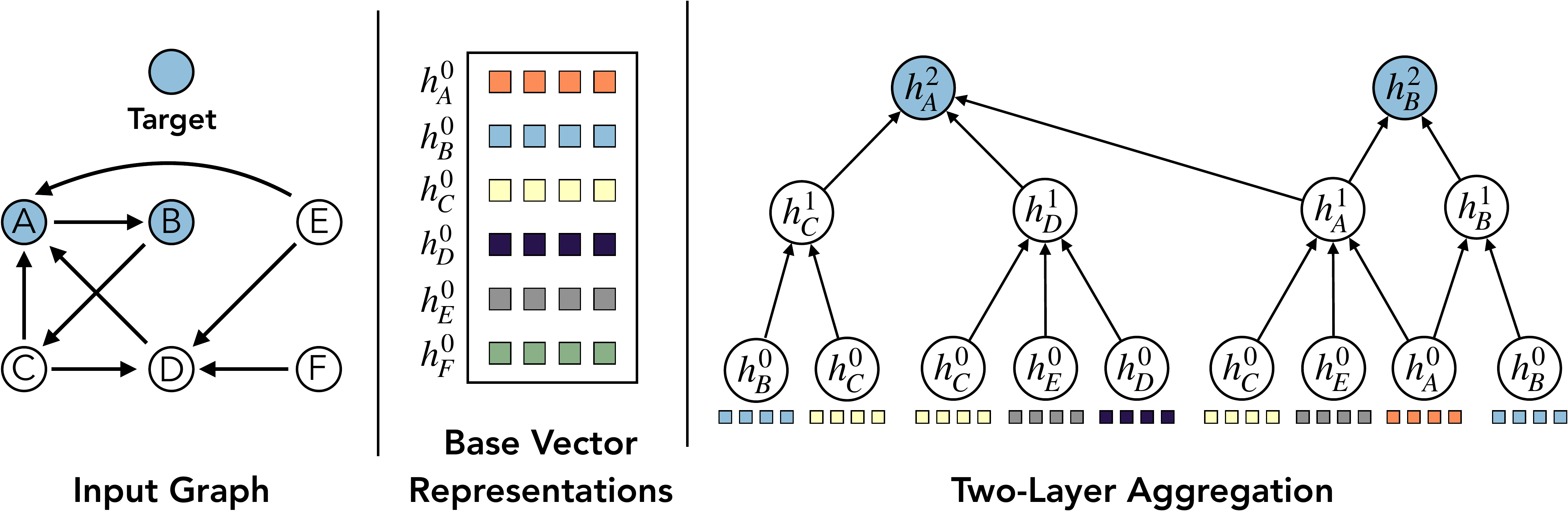}
  \vspace{-0.1in}
  \caption{Example two-layer GNN aggregation for nodes \{A, B\} using a sample of their two-hop incoming neighborhood.}
  \label{fig:background_fig}
  \vspace{-0.1in}
\end{figure}

To combine local information with neighborhood information, GNN layers aggregate the local representation of each node with the representations of its neighbors. Given a node $v$, the $k-$th layer of a multi-layer GNN represents node $v$ with a vector $h_v^k$ that is defined recursively as $h_v^k = AGG(h_v^{(k-1)}, \{h_u^{(k-1)} \colon u \in N_v\})$ where $AGG$ is an aggregation function with learnable parameters (e.g., a weighted sum) and $N_v$ denotes the nodes in the one-hop neighborhood of $v$; The $k$-th layer aggregates information from the $k$-hop neighborhood of node $v$. An example is shown in Figure~\ref{fig:background_fig}. To perform node classification, $h_v^k$ can be fed into a fully-connected and softmax layer, while for link prediction, $h_{v_1}^k$ and $h_{v_2}^k$ are given as input to a score function (typically referred to as a \emph{decoder}~\cite{chami2021machine}) to obtain the prediction.

GNN training is typically performed using mini-batch gradient descent. For node classification, training examples correspond to \emph{target nodes} together with the true class label for each node. For link prediction, training examples are pairs of target nodes with binary labels (i.e., edge or not). In a $k$-layer GNN the target nodes correspond to the nodes in the last ($k$-th) layer. To compute its output, a $k$-layer GNN requires that a mini batch include the base representation of all nodes in the $k$-hop neighborhood of the target nodes. However, $k$-hop neighborhoods can grow exponentially and the storage required for the base representations can exceed GPU memory if the entire neighborhood of each target node is used. For this reason, \emph{sampling} of the $k$-hop neighborhood, typically performed in CPU, is required.

Given a mini batch, a \emph{forward pass} is performed on the GPU to compute $h_v^k$ and the loss and gradients needed to update the model parameters. The updates for learned base representations are also transferred to the CPU and the corresponding entries of the lookup table are updated. The time for sampling the $k$-hop neighborhood can dominate the time for the forward pass, thus leading to GPU underutilization~\cite{kaler2022accelerating} and lower training throughput.

For large-scale graphs, the storage for the graph structure and the base representations of the nodes may exceed CPU memory (Table~\ref{tab:intro_table}). In this case, mini batch construction needs to be performed over a subgraph of the original graph that is loaded in main memory. This operation requires the original graph be split into \emph{partitions}. These partitions can either be loaded in different machines to perform distributed training or stored on disk and brought into main memory to perform training over the corresponding subgraph. \systemname uses disk-based training. We opt for \emph{pipelined disk-based training}~\cite{mohoney2021marius} where a buffer is used to hold partitions in main memory. In-buffer partitions are used to construct mini batches. To achieve high accuracy, we need to iterate over all available training examples. To this end, pipelined disk-based training requires swapping partitions between the buffer and disk according to a \emph{partition replacement policy}. For high throughput this policy should also minimize IO. Prior work on the Marius system~\cite{mohoney2021marius} introduced the SoTA greedy BETA policy which achieved near minimal IO when iterating over all graph edges. While BETA allowed Marius to train \textit{decoder-only} knowledge graph embedding models for link prediction, Marius does not support multi-hop neighborhood sampling, and thus cannot train $k$-layer GNNs for $k>0$, or node classification tasks. Moreover, when BETA is implemented in \systemname and applied to $k$-layer GNN training it leads to reduced model accuracy.

\section{Training GNNs in \systemname}
\label{sec:sys_overview}

\begin{figure*}[!t]
  \centering
  \includegraphics[width=1.0\textwidth]{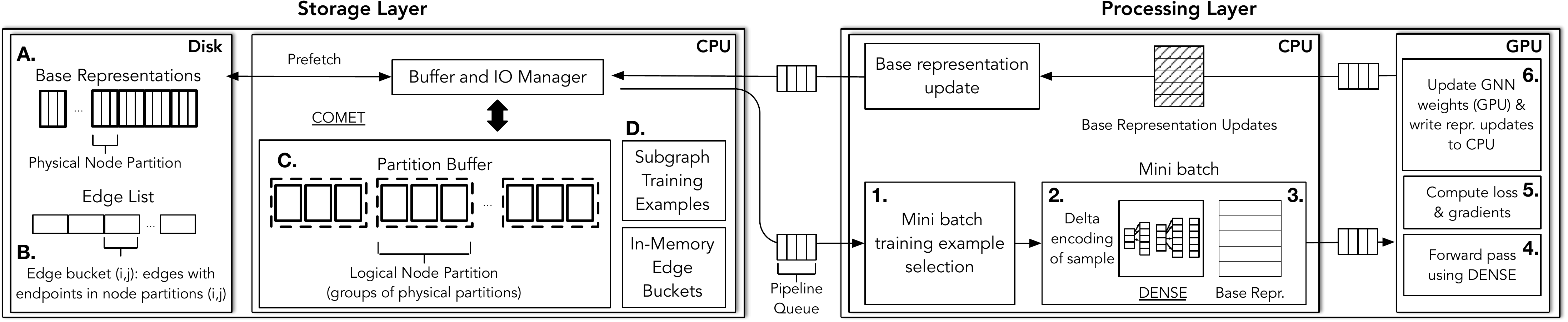}
  \vspace{-0.25in}
  \caption{\systemname System Diagram. The lifecycle of a mini batch consists of Steps 1-6. The storage layer replacement policy (e.g., \replacementpolicy) periodically updates graph partitions in memory during training (Steps A-D).}
  \label{fig:system_diagram}
  \vspace{-0.1in}
\end{figure*}

\systemname implements out-of-core pipelined training using two modules: 1) a processing layer and 2) a storage layer. Figure~\ref{fig:system_diagram} shows a diagram of GNN training in \systemname.

\systemname represents a graph as an edge list. In addition, base vector representations for nodes are stored sequentially in a lookup table split into $p$ \emph{physical partitions} on disk. The edge list is organized according to \emph{edge buckets}: Given a pair of partitions $(i, j)$, we define edge bucket $(i, j)$ to be the collection of all edges in the graph with a source node in partition $i$ and a destination node in partition $j$. Edges in each edge bucket are stored sequentially on disk. 

Training in \systemname proceeds in epochs. We start each epoch with all partitions on disk. We consider an epoch completed only when all training examples in the graph have been processed once. At the beginning of each epoch, \systemname groups the physical partitions into a collection of $l \leq p$ \emph{logical partitions}. The grouping is randomized and each physical partition is assigned to one logical partition. Grouping occurs without data movement: only an in-memory dictionary between logical and physical partitions is maintained. This two-level partitioning scheme is key to achieving high-accuracy GNNs (Section~\ref{subsec:disk_lp}). Given the logical partitions, let $S_i$ be a set of logical partitions such that all corresponding physical partitions fit in memory. \systemname constructs a sequence $S =\{S_1, S_2, \dots\}$ (to be consecutively loaded into memory during training) such that each training example appears in at least one $S_i$. We then use $S$ to obtain a sequence $X = \{X_1, X_2, \dots \}$. $X_i$ is a subset of training examples in $S_i$ such that when $S_i$ is in memory, all (and only) training examples from $X_i$ are used to generate mini batches. We describe the techniques used by \systemname to select $X_i$ in Section~\ref{sec:disk}.

Sequences $S$ and $X$ are generated in a task-aware manner. In the case of node classification, training examples are graph nodes and in the case of link prediction, they are pairs of nodes. Thus, for link prediction, pairs of node partitions that contain valid edges need to be accessed at the same time, while for node classification node partitions can be considered individually. For link prediction, \systemname uses the \replacementpolicy policy to generate $S$ and $X$. \replacementpolicy ensures that all valid edges of the graph, which correspond to training examples, will appear in at least one $S_i \in S$. \replacementpolicy also allows \systemname to minimize disk-to-CPU IO while maximizing accuracy for disk-based training. We describe \replacementpolicy in Section~\ref{subsec:disk_lp}. For node classification, \systemname uses a simple policy that caches all nodes used as training examples in memory to achieve both high accuracy and throughput (Section~\ref{subsec:disk_nc}). Finally, the above design also allows \systemname to support training with the full graph in memory: $S_1$ contains the whole graph and $X_1$ contains all training examples.

To complete one epoch (for either task), the storage layer in \systemname uses sequence $S$ to determine which node partitions and edge buckets to bring into the buffer and in what order. \systemname uses a buffer with capacity of $c$ physical node partitions. When the set of $c$ physical partitions in $S_i$ are placed in the buffer, all $c^2$ pairwise edge buckets are also loaded into memory. After loading $S_i$, the storage layer passes $X_i$ to the processing layer. 

The processing layer generates mini batches from $X_i$ in a random order. \systemname performs multi-hop neighborhood sampling using the \datastructure data structure to construct a mini batch (Section~\ref{sec:mini_batch}). To improve throughput, \datastructure allows \systemname to minimize redundant computation during sampling by reusing neighborhood samples required as input to different GNN layers. Neighborhood sampling is performed only over graph nodes and edges in main memory. \datastructure and the corresponding base vector representations are then transferred to the GPU to complete processing of the mini batch. \datastructure is co-designed such that the forward pass for the GNN is computed using kernels that are optimized for dense linear algebra operations. After the forward pass, \systemname computes the loss and gradients. We update GNN parameters on the GPU and if applicable, updates to learnable base vector representations are transferred back to CPU memory and used to update the node representations in the partition buffer. As done in prior work~\cite{mohoney2021marius}, we perform all data transfers in a pipelined manner as shown in Figure~\ref{fig:system_diagram}. 

After training completes on the mini batches generated from $X_i$, the storage layer updates the partitions in the buffer from $S_i$ to $S_{i+1}$ by swapping the necessary logical partitions between disk and CPU memory (one or more physical partitions) together with the corresponding edge buckets. This process repeats until the epoch is completed.

\section{The \datastructure Data Structure}
\label{sec:mini_batch}
\datastructure is designed to 1) minimize redundant computation in multi-hop sampling and 2) to allow for efficient GNN forward passes using dense kernels for linear algebra operations.

\subsection{Multi-hop Neighborhood Samples With \datastructure}
\label{subsec:dense_sampling}
To compute the representation of a set of target nodes after $k$ GNN layers, we need to sample their $k$-hop neighborhood. For example, in Figure~\ref{fig:background_fig}, we showed how to compute the representation of target node $A$ after two layers. To compute $h^2_A$ we used $h^1_C$ and $h^1_D$ by sampling $C$ and $D$ from $A$'s one-hop neighborhood. Computing $h^2_A$, however, uses $h^1_A$ which requires sampling the one-hop neighborhood of node $A$ again. Performing these two one-hop sampling operations independently requires traversing the one-hop neighborhood of node $A$ multiple times, introducing redundant computation.

\begin{figure}
  \centering
  \includegraphics[width=0.41\textwidth]{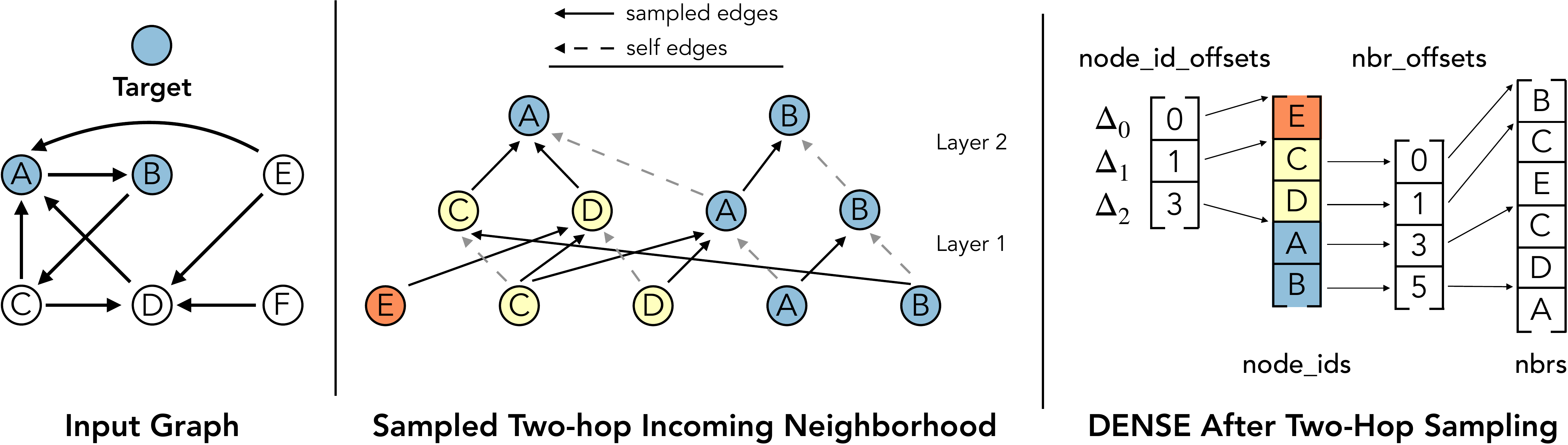}
  \caption{Example two-hop neighborhood sample for target nodes $\{A, B\}$ and the corresponding \datastructure data structure.}
  \label{fig:data_structure}
\end{figure}

In \systemname, we sample one-hop neighbors for each node in the $k$-hop neighborhood \textit{only once}. We take advantage of the fact that $k$-hop sampling is recursive: we can construct a sample of the $(i+1)$-hop neighborhood for a set of target nodes by sampling the one-hop neighbors of all nodes $N$ in the $i$-hop neighborhood. However, we may have previously sampled one-hop neighbors for some nodes $P \subseteq N$. In \systemname, we use this property to construct the $(i+1)$-hop neighborhood by sampling one-hop neighbors only for the nodes $N \setminus P$. We define these nodes to be $\Delta_{k-i}$. The rest of the $(i+1)$-hop neighborhood is completed by reusing the previous one-hop samples for nodes in $P$. To track the nodes for which one-hop samples are required at each iteration, \systemname uses the \datastructure data structure. \datastructure is constructed by stacking the $k+1$ $\Delta$'s and the corresponding one-hop neighbors. We show an example for the two-hop neighborhood of the target nodes $\{A, B\}$ in Figure~\ref{fig:data_structure}. \datastructure consists of three $\Delta$'s: $\Delta_2 = \{A, B\}$, $\Delta_1 = \{C, D\}$, $\Delta_0 = \{E\}$. Notice that unlike in Figure~\ref{fig:background_fig}, here, the sampled one-hop neighbors of $A$, $\{C, D\}$ are used in both GNN layers.

\begin{algorithm}[!t] \footnotesize
    \SetInd{.25em}{1em}
    \SetAlgoLined
    \DontPrintSemicolon
    \KwIn{target\_nodes: unique node IDs for $k$-hop sampling; $\;\;$
    fanouts: max \# of neighbors to sample per hop}
    
    node\_id\_offsets = $[0]$; node\_ids = target\_nodes
    
    nbr\_offsets = $[]$; nbrs = $[]$; $\Delta_k$ = target\_nodes
    
    \For{$i$ in $[k \dots 1]$}{
    
        $\Delta_i$\_nbrs, $\Delta_i$\_offsets = \texttt{oneHopSample}$(\Delta_i$, fanouts$[i])$
        
        nbr\_offsets = \texttt{cat}$(\Delta_i$\_offsets, nbr\_offsets + \texttt{len}$(\Delta_i$\_nbrs$))$
        
        nbrs = \texttt{cat}$(\Delta_i$\_nbrs, nbrs$)$

        $\Delta_{i-1}$ = \texttt{computeNextDelta}$(\Delta_i$\_nbrs, node\_ids$)$
        
        node\_id\_offsets = \texttt{cat}$([0]$, node\_id\_offsets + \texttt{len}$(\Delta_{i-1}))$
        
        node\_ids = \texttt{cat}$(\Delta_{i-1}$, node\_ids$)$
    }
    \Return DENSE(node\_id\_offsets, node\_ids, nbr\_offsets, nbrs)
    \caption{Multi-hop Neighborhood Sampling}
    \label{alg:data_structure_creation}
\end{algorithm}

\datastructure is built using four arrays: 1) \texttt{node\_ids} contains all graph node IDs involved in the sample, 2) \texttt{nbrs} contains the sampled one-hop neighbors for nodes in \texttt{node\_ids}, 3) \texttt{nbr\_offsets} identifies where the neighbors for each node ID start in \texttt{nbrs}, and 4) \texttt{node\_id\_offsets} identifies groups of node IDs in \texttt{node\_ids} corresponding to each $\Delta$. Given a set of target node IDs and a $k$-layer GNN, \systemname samples the $k$-hop neighborhood for each target node and creates the four arrays in \datastructure according to Algorithm~\ref{alg:data_structure_creation}. We define the target nodes to be $\Delta_k$ and initialize each array in \datastructure as shown (Line 1-2). Sampling then proceeds for $k$ rounds (Line 3). Each iteration $i \in [k \dots 1]$ starts by sampling the one-hop neighbors for the nodes in $\Delta_i$ (Line 4). Given the set of nodes $\Delta_i$ and a maximum number of neighbors to sample per node $f$ (the layer \textit{fanout}), one-hop sampling returns up to $f$ neighbors for each node $j \in \Delta_i$ as a list \texttt{$\Delta_i$\_nbrs} with the neighbors for each node $j$ sequential starting at the offset given by \texttt{$\Delta_i$\_offsets}. When a node has more than $f$ neighbors, only $f$ will be sampled, but if a node has less than $f$ neighbors, all neighbors will be returned. 

\systemname performs one-hop sampling using CPU multi-threading. We store two sorted versions of the in-memory edge list containing all edges between the node partitions currently in memory: 1) sorted in ascending order of source node ID, and 2) sorted in ascending order of destination node ID. We create an array that, for each node ID in memory, stores the offsets corresponding to its outgoing and incoming edges in each of the two edge lists. Given these structures, we can sample incoming and outgoing edges for any set of nodes in parallel using all available CPU threads.

Given the one-hop neighbors for $\Delta_i$, the next step in Algorithm~\ref{alg:data_structure_creation} is to stack these one-hop samples on the existing arrays in \datastructure (Line 5-6). At this point, \systemname computes $\Delta_{i-1}$ as the unique nodes in \texttt{$\Delta_i$\_nbrs} that do not appear in the \datastructure \texttt{node\_ids} array (Line 7). As with one-hop sampling, \systemname computes $\Delta_{i-1}$ using multi-threading on the CPU. The nodes in $\Delta_{i-1}$ are then added to \datastructure(Line 8, 9). Multi-hop sampling completes after adding $\Delta_0$ to \datastructure (neighbors are not needed for $\Delta_0$). 

While one-hop sample reuse in \datastructure allows \systemname to minimize redundant computation in multi-hop sampling, it also leads to $k$-hop neighborhoods with two noteworthy differences compared to existing sampling algorithms. First, sample reuse implies that the resulting neighborhood fanouts (max number of neighbors per node per hop) are not guaranteed to match the requested fanouts (input to Algorithm~\ref{alg:data_structure_creation}). Instead, the fanout for a node $j$ at each hop is equal to the fanout requested for the first hop which required neighbors of $j$. 
As such, in the common case where GNN fanouts are requested according to a decreasing sequence away from the target nodes, \datastructure provides at least as many neighbors as requested for a node $j$ at each layer. Second, sample reuse in \datastructure reduces randomness in multi-hop neighborhoods compared to existing algorithms by preventing different subsets of the one-hop neighbors for a given node from existing in the same multi-hop neighborhood. We discuss the implication of this reduced randomness on the accuracy of GNN training in Section~\ref{subsec:e2e_comp} but find that training with \datastructure can reach comparable accuracy to existing systems.

\subsection{Forward Pass Computation With \datastructure}
After sampling is completed, \systemname transfers \datastructure to the GPU so it can be used to compute the GNN forward pass. We also transfer an array $H^0$ containing the base representations for each node ID in the \datastructure \texttt{node\_ids} array. On the GPU we create and add a fifth array to \datastructure called \texttt{repr\_map} that stores the index in $H^0$ containing the base representation for each node ID in the \datastructure \texttt{nbrs} array. 

To complete a forward pass over a $k$-layer GNN, we iterate over each layer $i \in [1 \dots k]$ and perform the next two steps: 

\vspace{2pt}\noindent (Step 1) We compute the output $H^i$ of layer $i$. Given \datastructure and the representations in $H^{i-1}$, the output of layer $i$ is the representation for all nodes in the \datastructure \texttt{node\_ids} array which occur after \texttt{node\_id\_offsets}$[1]$. For example, in Figure~\ref{fig:data_structure}, the output of the first GNN layer is $h^1$ for the nodes $\{C, D, A, B\}$. The output representations are computed according to the $i^{th}$ GNN layer's aggregation function.

\vspace{2pt}\noindent (Step 2) We update \datastructure on the GPU as shown in Algorithm~\ref{alg:update_dense}: We remove nodes and their one-hop neighbors that are no longer needed for subsequent layers. This step ensures that the output nodes from this iteration, which will be used as input to compute the representations of the nodes in Step 1 for the next iteration, correspond to all nodes in the \datastructure \texttt{node\_ids} array. This property allows us to use the same implementation across GNN layers. In Figure~\ref{fig:data_structure}, node $E$ and the neighbors of $\{C, D\}$ are not needed after layer one.

\datastructure allows \systemname to use optimized dense GPU kernels for the linear algebra operations in each GNN layer. For example, in Algorithm~\ref{alg:gnn_layer}, we show how to use \datastructure and $H^{k-1}$ to compute the output of the $k^{th}$ GNN layer $H^k$ for the GNN aggregation: $h^{(l+1)}_i = h^{(l)}_i + \sum_{j \in Nbrs_i} h^{(l)}_j$. We first use the \texttt{repr\_map} array in \datastructure to select the node representations for all neighbors in the \texttt{nbrs} array (Line 1). Neighborhood aggregation can then be performed using a dense segment sum that is well suited for parallelization on GPU hardware (Line 2): Recall that the one-hop neighbors for the nodes in \datastructure are stored sequentially and separated by the \texttt{nbr\_offsets} array. After selecting the neighbor representations in Line 1, the representations for the one-hop neighbors of each node will also be sequential in GPU memory. Thus, neighborhood aggregation for each node consists of adding a set of sequential vectors and all nodes can aggregate in parallel. The last step to compute the layer output is to combine the aggregated neighbor representations for each node with their own representations (Lines 3-4).

\begin{algorithm}[!t] \footnotesize
    \SetInd{.25em}{1em}
    \SetAlgoLined
    \DontPrintSemicolon
    \KwIn{node\_id\_offsets, node\_ids, nbr\_offsets, nbrs, repr\_map}
    
    $\Delta_{i-1}$ = node\_ids$[$: node\_id\_offsets$[1]]$
    
    $\Delta_{i}$ = node\_ids$[$node\_id\_offsets$[1]$ : node\_id\_offsets$[2]]$
    
    $\Delta_i$\_nbrs = nbrs$[$: nbr\_offsets$[$\texttt{len}$(\Delta_i)]]$

    nbrs = nbrs$[$\texttt{len}$(\Delta_i$\_nbrs$)$:$]$
    
    repr\_map = repr\_map$[$\texttt{len}$(\Delta_i$\_nbrs$)$:$]$ - \texttt{len}$(\Delta_{i-1})$
    
    nbr\_offsets = nbr\_offsets$[$\texttt{len}$(\Delta_i)$:$]$ - \texttt{len}$(\Delta_i$\_nbrs$)$
    
    node\_ids = node\_ids$[$node\_id\_offsets$[1]$:$]$
    
    node\_id\_offsets = node\_id\_offsets$[1$:$]$ - \texttt{len}$(\Delta_{i-1})$
    
    \Return node\_id\_offsets, node\_ids, nbr\_offsets, nbrs, repr\_map
    \caption{On GPU \datastructure Update After Layer $i$}
    \label{alg:update_dense}
\end{algorithm}

\begin{algorithm}[!t] \footnotesize
    \SetInd{.25em}{1em}
    \SetAlgoLined
    \DontPrintSemicolon
    \KwIn{DENSE; $H^{k-1}$: layer input vector representations}
    
    nbr\_repr = $H^{k-1}$.\texttt{index\_select}(DENSE.repr\_map)
    
    nbr\_aggr = \texttt{segment\_sum}$($nbr\_repr, DENSE.nbr\_offsets$)$
    
    self\_repr = $H^{k-1}[$DENSE.node\_id\_offsets$[1]$ :$]$
    
    $H^k$ = nbr\_aggr + self\_repr

    \Return $H^k$
    \caption{$k^{th}$ GNN Layer Additive Aggregation}
    \label{alg:gnn_layer}
\end{algorithm}

\section{Policies for Disk-Based GNN Training}
\label{sec:disk}
We discuss the partition replacement policies used by \systemname for disk-based training. As described in Section~\ref{sec:sys_overview}, \systemname uses different policies for link prediction and node classification. We discuss each in turn.

\subsection{Policies for Link Prediction}
\label{subsec:disk_lp}
\systemname uses the \replacementpolicy policy to maximize throughput while maintaining high accuracy when training link prediction models. Before we introduce \replacementpolicy, we discuss why policies that only optimize for throughput lead to biased training and hence harm the accuracy of the learned models.

\begin{figure}[!t]
  \centering
  \includegraphics[width=0.40\textwidth]{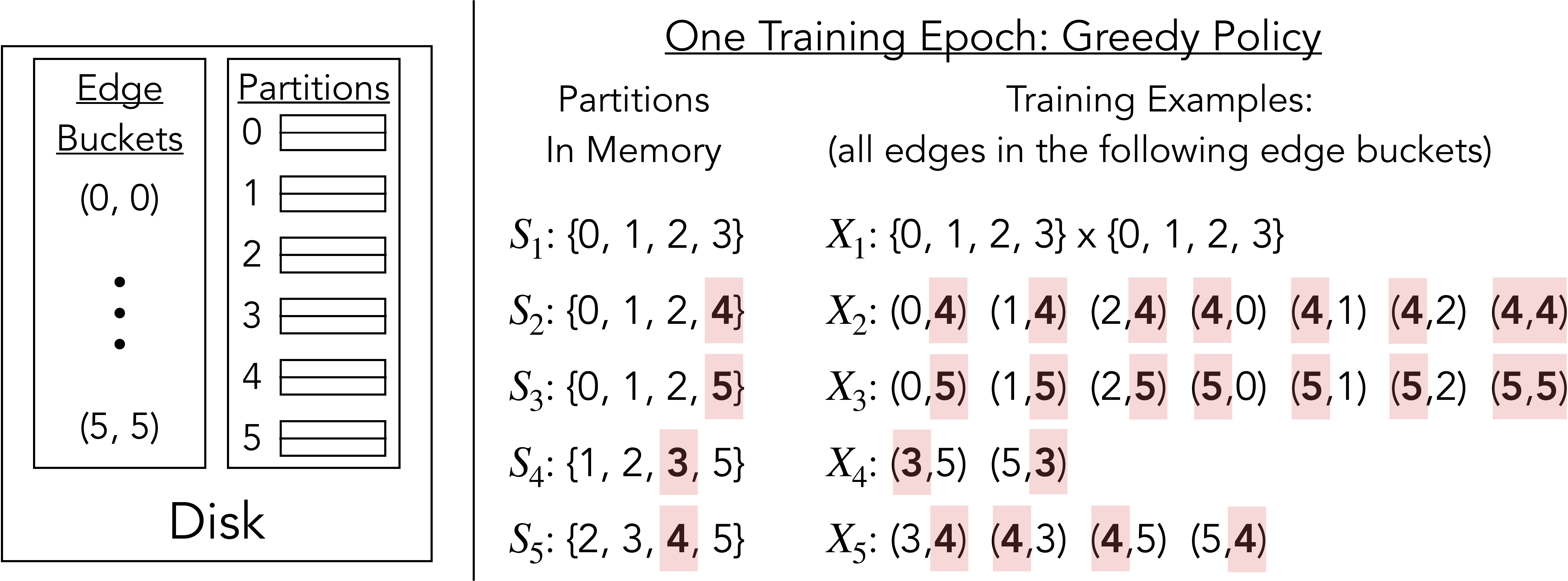}
  \vspace{-0.1in}
  \caption{Greedy sequence of partitions in memory $S$ and training examples $X$ that are correlated. E.g., the examples in $X_2$ all come from edge buckets containing partition four.}
  \label{fig:correlated_examples}
  \vspace{-0.05in}
\end{figure}

Greedy policies (e.g., BETA~\cite{mohoney2021marius}) that focus on minimizing IO for high throughput produce correlated training examples that bias learning and lead to low model accuracy. Recall that we define $S = \{S_1, S_2, \dots\}$ to be the sequence of partition sets which will be loaded into memory during one epoch and $X = \{X_1, X_2, \dots\}$ to be the sequence of training examples used to generate mini batches for each $S_i \in S$ (Section~\ref{sec:sys_overview}). To minimize IO, greedy policies swap partitions between $S_i$ and $S_{i+1}$ such that the new partitions brought into memory maximize the number of new training examples that can be generated from the in-memory graph. For example, the BETA policy minimizes IO by bringing one new physical partition $p^*$ in memory to obtain $S_{i+1}$ and uses the edges (training examples) that correspond to node pairs formed by combining $p^*$ with all other partitions in memory to construct $X_{i+1}$. This process makes all training examples in $X_{i+1}$ be correlated: they all have one endpoint in the new partition $p^*$. We show an example of this problem in Figure~\ref{fig:correlated_examples}. As highlighted in the introduction, performing training over correlated examples reduces randomness in the order edges are processed each epoch and conflicts with the independently distributed assumption of ML training data. In Section~\ref{sec:eval_disk_pol}, we show that using a greedy policy leads to accuracy degradation compared to training with the full graph in memory.

\begin{figure}[!t]
  \centering
  \includegraphics[width=0.44\textwidth]{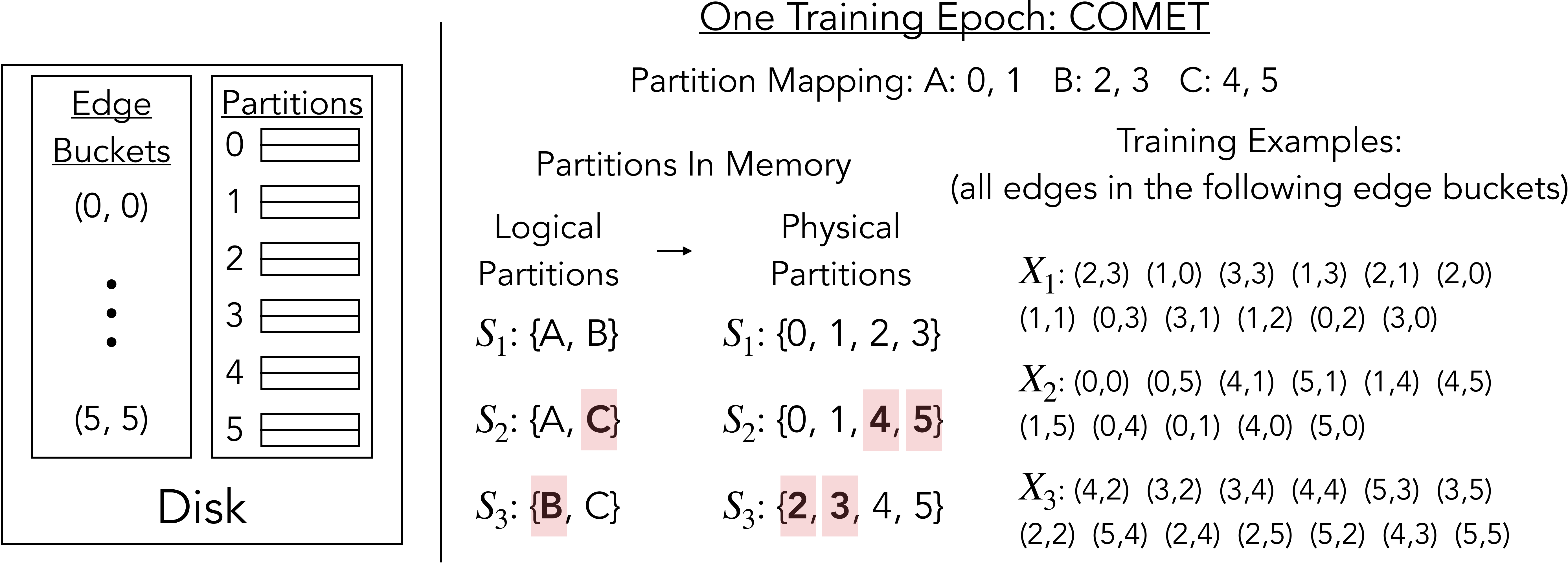}
  \vspace{-0.1in}
  \caption{Partition and training example sequences generated by \replacementpolicy to minimize training example correlation. }
  \label{fig:comet}
  \vspace{-0.125in}
\end{figure}

\replacementpolicy addresses the above shortcoming by introducing randomness in the order that training examples are processed each epoch while simultaneously minimizing IO. To increase randomness, we design \replacementpolicy around two mechanisms: 1) a two-level (logical and physical) partitioning scheme and 2) randomized generation of training examples. Following the architecture from Section~\ref{sec:sys_overview}, the first mechanism generates the sequence of partition sets $S$ for one epoch and the second mechanism generates the sequence of training examples $X$. 

To decouple data storage and access from data transfer, \replacementpolicy uses physical partitions on disk but transfers groups of physical partitions---called logical partitions---between disk and CPU memory. At the beginning of each epoch, physical partitions are randomly grouped into logical partitions (without data movement) (Section~\ref{sec:sys_overview}). \replacementpolicy then generates $S = \{S_1, S_2, \dots\}$ by greedily swapping one logical partition between $S_i$ and $S_{i+1}$ such that all pairs of partitions (and thus pairs of nodes) appear in at least one $S_i$ with minimal IO. By utilizing logical partitions, \replacementpolicy allows \systemname to improve randomness by utilizing small physical partitions---which fix fewer nodes together in a partition for the whole training process---yet also use large logical partitions to increase turnover rate of graph data between each $S_i$. In Section~\ref{sec:hyperparameters}, we analyze how to best set the number of physical and logical partitions to simultaneously minimize IO and maximize accuracy. 

Instead of using logical partitions to increase randomness in the sequence of partitions sets $S$, an alternative design would be to create a new greedy algorithm to minimize IO while considering multiple physical partition swaps at once. In \systemname we opt to use logical partitions for the following reasons. First, by swapping one logical partition at a time \systemname can utilize existing one-swap greedy algorithms that have been shown to minimize total epoch IO near the theoretical lower bound~\cite{mohoney2021marius}. Thus a multi-swap greedy algorithm can at best provide little IO benefit. Finally, allowing for multiple swaps at once exponentially increases the number of swap choices to consider between each set of partitions $S_i$ and $S_{i+1}$, making it challenging to develop an efficient multi-swap algorithm to generate $S$. Thus, we focus on using a two-level partitioning scheme in \systemname.

Beyond introducing randomness in $S$, \replacementpolicy also injects randomness in the sequence of training examples $X$ used to create mini batches. Given that $S$ is generated at the beginning of each epoch, \systemname performs the following optimization: For each pair of partitions $(i, j)$ in the graph, \systemname identifies all partition sets $S_{(i,j)} \subseteq S$ that contain both $i$ and $j$. \replacementpolicy then picks one $S_*$ at random from $S_{(i, j)}$ and assigns the training examples corresponding to pairs of nodes between these two partitions---the edges in edge bucket $(i, j)$---to $X_*$. This random assignment allows for the deferred processing of training examples rather than greedily processing all new examples immediately upon arrival in CPU memory. Beyond shuffling training examples, this deferred execution scheme also balances the workload across each $X_i$---each $X_i$ contains in expectation the same number of training examples. When prefetching is used to mask the IO latency required to load $S_{i+1}$ during mini-batch training on $S_i$, balanced workloads enable consistent overlapping of IO and compute. In contrast, greedy policies generate unbalanced workloads where some $X_i$ contain very few training examples (e.g., Figure~\ref{fig:correlated_examples}). For these cases, training on $X_i$ completes before $S_{i+1}$ is loaded leading to IO bottlenecks.

\subsection{Policies for Node Classification}
\label{subsec:disk_nc}
For node classification we find that a simple replacement policy is sufficient to maximize throughput without harming accuracy. To iterate over all training examples during each epoch, we require that all labeled graph nodes in the training set---called the \textit{training nodes}---appear in memory at least once (in at least one $S_i$). We find that in large-scale graphs~\cite{hu2021ogblsc, hu2020ogb}, it is often the case that the training nodes make up only one to ten percent of all graph vertices. As such, the base representations for the training nodes can fit in CPU memory, even when the storage overhead for the full graph is many times larger. When this observation holds, for disk-based node classification, \systemname performs static caching of the training nodes and their base representations in CPU memory. While prior works have also used static caching for GNN training~\cite{pagraph, yang2022gnnlab}, these approaches focus on caching hot vertices in GPU memory to minimize CPU to GPU data transfer rather than caching training examples in CPU memory to minimize disk to CPU transfers.

More specifically, we perform disk-based node classification as follows: Rather than randomly partitioning the graph, we assign all training nodes sequentially to the first $k$ physical partitions. Non-training nodes are assigned to physical partitions randomly as before. We generate one set of partitions to be loaded into memory each epoch $S = \{S_0\}$. $S_0$ contains the $k$ partitions with training nodes together with $c-k$ other randomly chosen physical partitions (buffer capacity $c$). By construction, all training nodes are assigned to create mini batches in $X_0$. This policy leads to zero partition swaps (IO) during an epoch (IO does occur between epochs), but assumes that $k$ is less than $c$ (all training examples can fit in CPU memory). If $k \geq c$, \systemname uses random partitioning and \replacementpolicy but generates $S$ as follows: replace a random logical partition in memory with a random one from disk that has not appeared in memory until all partitions have appeared in the buffer. In the future, we plan to study how this approach compares with other schemes~\cite{liu2021bgl} for dynamically caching training examples in CPU memory.

\section{Auto-tuning Rules For Disk Training}
\label{sec:hyperparameters}

\systemname provides auto-tuning for 1) the number of physical partitions $p$, 2) the number of logical partitions $l$, and 3) the buffer capacity $c$ to minimize training time and maximize model accuracy out of the box. We focus on $p$ and $l$ since maximizing $c$ best approximates training with the full graph in memory and thus leads to better runtime and accuracy. We first connect $p$ and $l$ to model accuracy, then focus on their effect on runtime, and conclude by using this information to present auto-tuning rules (for $p$, $l$, and $c$).

\begin{figure*}
  \centering
  \begin{subfigure}[t]{0.275\textwidth}
     \centering
     \includegraphics[width=\textwidth]{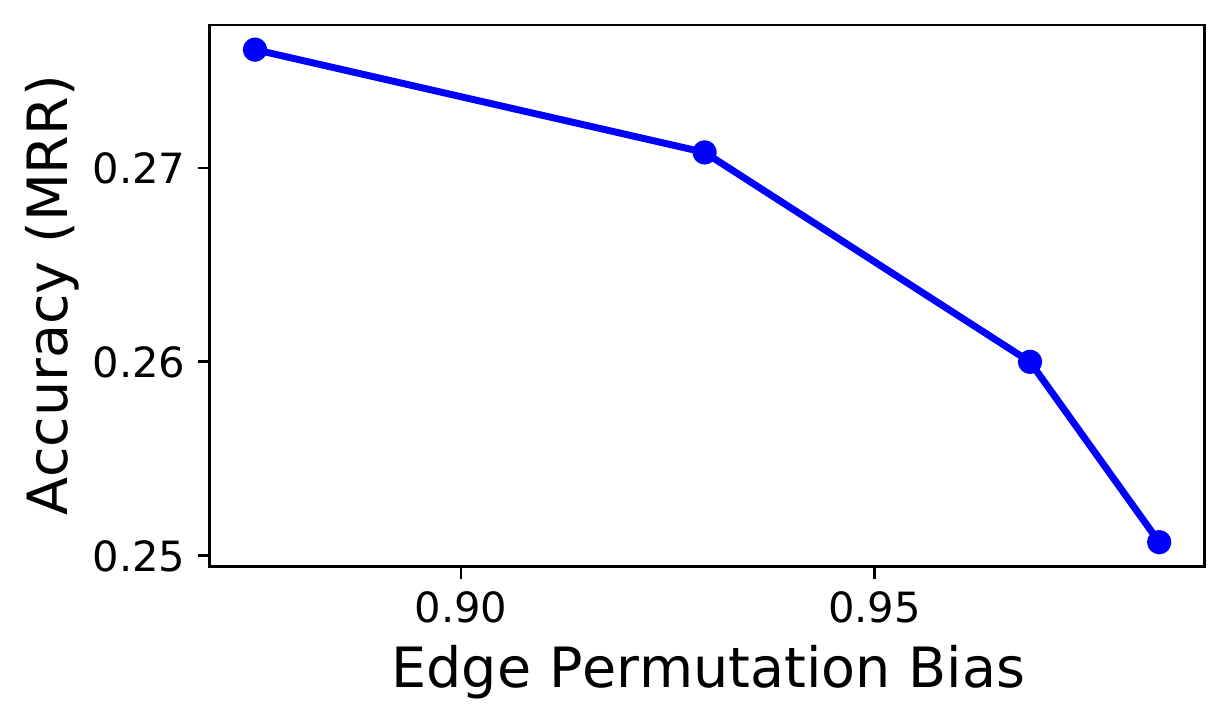}
     \caption{Model Accuracy vs. Bias $B$}
     \label{fig:bias_vs_accuracy}
  \end{subfigure}
  \hspace{0.02\textwidth}
  \begin{subfigure}[t]{0.38\textwidth}
     \centering
     \includegraphics[width=\textwidth]{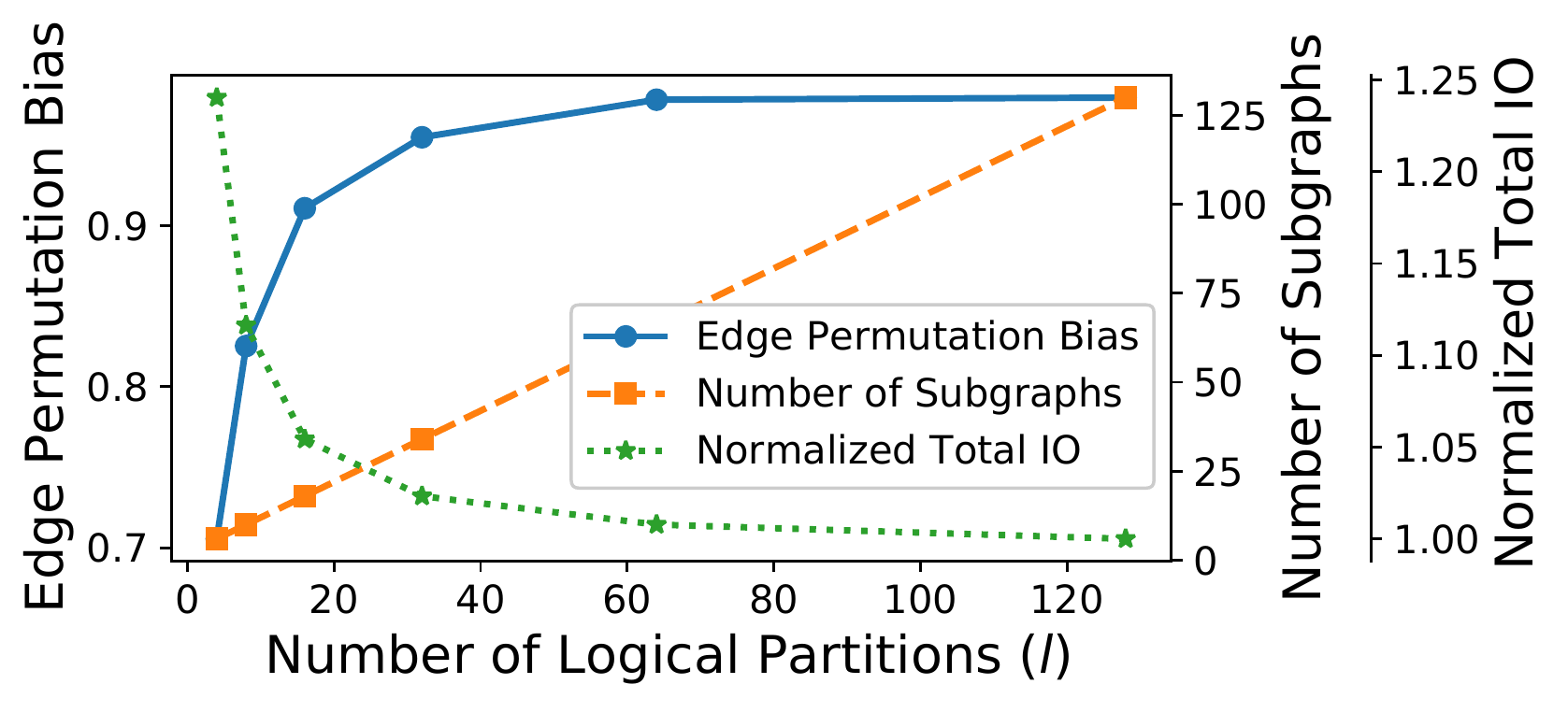}
     \caption{Effect of logical partitions}
     \label{fig:logical_partitions}
  \end{subfigure}
  \hspace{0.02\textwidth}
  \begin{subfigure}[t]{0.275\textwidth}
     \centering
     \includegraphics[width=\textwidth]{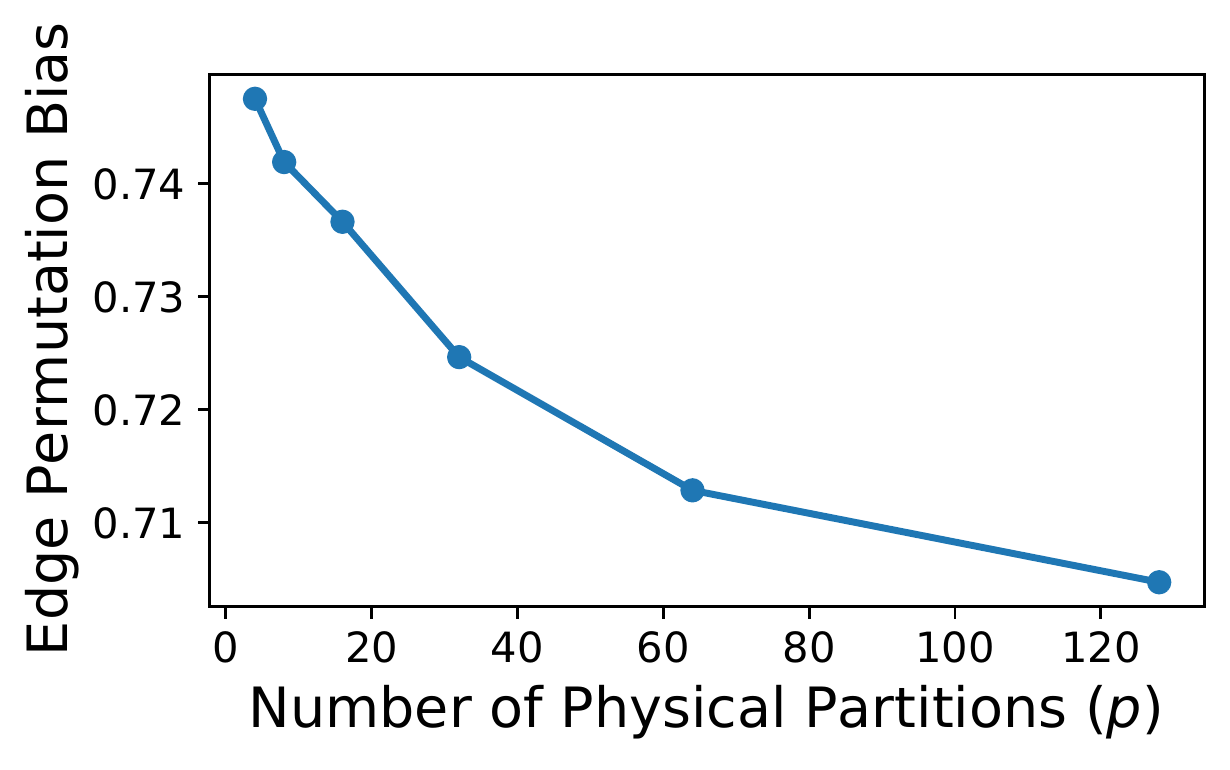}
     \caption{Effect of physical partitions}
     \label{fig:physical_partitions}
  \end{subfigure}
  \vspace{-0.1in}
  \caption{Empirical measurements on the effect of \replacementpolicy parameters using GraphSage on FB15k-237.}
  \vspace{-0.1in}
  \label{fig:convergence}
\end{figure*}

\vspace{0.025in}
\newparagraph{Effect of $p$ and $l$ on Accuracy}
To study the effect $p$ and $l$ have on model accuracy, we introduce a proxy metric which we term the \textit{Edge Permutation Bias} $B$. Recall that model quality can degrade if training consecutively iterates over correlated examples (Section~\ref{subsec:disk_lp}). This problem is general to ML workloads~\cite{pmlr-v97-haochen19a, NEURIPS2020_42299f06, hofmann2015variance}. We design $B$ to capture the extent to which the sequence of training examples generated by \replacementpolicy exhibits this phenomenon. Figure~\ref{fig:bias_vs_accuracy} illustrates the dependency between $B$ and model accuracy. The depicted results correspond to empirical measurements over a benchmark model (GraphSage~\cite{graphsage}) and dataset (FB15k-237~\cite{fb15k237}). We find the same behavior to hold across settings.

We now define $B$. Let $X = \{X_1 \dots X_n\}$ be the sequence of \textit{edge bucket sets} $X_i$ assigned as training examples for each partition set $S_i$ (Section~\ref{subsec:disk_lp}). If $X_i$ contains edges that focus on a small subset of nodes, then we have the undesired correlation described above. We empirically measure this occurrence as follows: Let $V$ be the set of nodes in the graph. For each node $v \in V$ we keep a tally $t_i^v$ as we iterate over $X$ which measures how many edges we have seen containing this node after each $X_i$. Tallies are cumulative and we assume a uniform degree distribution. We normalize such that $t_n^v = 1$. This implies $t_i^v \in [0, 1]$. After each $X_i$ we calculate $d_i = max_{v1, v2 \in V} (t_i^{v1} - t_i^{v2})$. Given this, $B = max_i d_i \in [0, 1]$.

We are interested in how evenly the tallies are incremented during a training epoch. Biased assignments will lead to processing many edges for a subset of nodes at once while ignoring the remaining graph vertices. This leads to high variance in model gradients across the epoch. With this in mind, $B = z$ means that $z+a$ percent of the edges containing a certain node have been processed before $a$ percent of the edges of another node have been processed (for $a \in [0, 1-z]$).

Figure~\ref{fig:physical_partitions} shows that $B$ decreases with increasing physical partitions. The observed trends can be characterized by the equation $B = \mathcal{O} (p^{-\alpha_1})$ for some constant $\alpha_1 > 1$. Figure~\ref{fig:logical_partitions} shows the effect of the number of logical partitions on $B$. Decreasing the number of logical partitions decreases the Edge Permutation Bias roughly according to $B = \mathcal{O} (l^{\alpha_2})$ for $0 < \alpha_2 < 1$. While we do not provide a closed-form characterization of $B$ as a function of $p$ and $l$ we find that the aforementioned trends hold across datasets. Moreover, describing the limiting behavior of $B$ with respect to $p$ and $l$ suffices to obtain a concrete methodology for optimizing these hyperparameters (described below).

\vspace{0.05in}
\newparagraph{Effect of $p$ and $l$ on Training Time}
We now focus on how $p$ and $l$ affect the per-epoch runtime $T$. To do so, we analyze three metrics that influence training time: 1) the total IO in terms of bytes transferred from disk to CPU memory ($IO$), 2) the number of partition sets generated per epoch ($|S|$), and 3) the smallest size of disk reads ($R$) in bytes. As Quantities 1 and 2 increase, the total training time increases---recall from Section~\ref{subsec:dense_sampling} that preparing each $S_i$ for training requires creating single-hop sampling data structures---while a decrease in Quantity 3 leads to an increased runtime. 

By construction in \replacementpolicy, the number of logical partitions affects Quantities 1 and 2. Figure~\ref{fig:logical_partitions} shows that as $l$ increases the total IO decreases and the number of partition sets per-epoch increases. The limiting behavior of $IO$ and $|S|$ with respect to $l$ is: $IO = \mathcal{O} (l^{-\alpha_3})$ for $\alpha_3 > 1$ and $|S|$ = $\mathcal{O}(l)$. For the purposes of this work, we assume that the training time is dominated by the number of partition sets ($|S|$) per epoch instead of the total IO for two reasons: First, the relative difference between the best and worst $IO$ is usually only between 5-25 percent and second, prefetching can overlap $IO$ with compute. Thus, we take the training time $T = \mathcal{O}(l)$. 

The training time $T$ is also affected by the number of physical partitions $p$ through Quantity 3. As $p$ increases, the size of each partition decreases linearly and the size of each edge bucket decreases quadratically---the smaller of these quantities is the smallest disk read size $R$. As a result, disk access transitions from large sequential reads/writes to small random reads/writes with increasing $p$. Given the hardware constraints of block storage, the latter can become a bottleneck, particularly when read sizes $R$ are less than the disk block size $D$. Thus, we model the affect of $p$ on training time according to $T = \mathcal{O}(1)$ for $p \leq \alpha_4$ and $\mathcal{O}(p)$ for $p > \alpha_4$, with $\alpha_4$ representing the number of partitions which cause the smallest disk reads to equal the block size $D$.

\vspace{0.05in}
\newparagraph{Methodology for Setting Hyperparameters}
Given the effect of $p$ and $l$ on accuracy and training time described above, together with the desire to maximize the buffer capacity $c$, we now present rules for setting the \replacementpolicy hyperparameters. We assume a graph $G = (V, E)$, that the base vector representation of each node is of dimension $d$, that CPU memory is of capacity $CPU$ bytes, and that the disk block size is $D$. First, we calculate the total overhead of storing all node representations as $NO = |V| * d * 4$ bytes (using floating point numbers). Likewise, the edge overhead $EO$ can be calculated from $|E|$ and the number of bytes per edge. Then, the overhead of each node partition is $PO = NO/p$ and the expected size of each edge bucket is $EBO = EO / p^2$.

With the above definitions, $p$ affects the Edge Permutation Bias $B$ and training time $T$ as follows: $B = \mathcal{O} (p^{-\alpha_1})$ for some constant $\alpha_1 > 1$ and $T = \mathcal{O}(1)$ for $p \leq \alpha_4$ and $\mathcal{O}(p)$ for $p > \alpha_4$ with $\alpha_4 = min(NO/D, \sqrt{EO/D})$. Thus, to minimize $B$ without increasing $T$, we set $p = \alpha_4$. We then maximize $c$ such that $c*PO + 2 * c^2 * EBO + F < CPU$. The edge term is multiplied by two because \systemname utilizes two sorted versions of the edge list (Section~\ref{subsec:dense_sampling}) and we leave some extra CPU space for working memory (fudge factor $F$). Finally, $B$ and $T$ are affected by the number of logical partitions according to $B = \mathcal{O} (l^{\alpha_2})$ for $0 < \alpha_2 < 1$ and $T = \mathcal{O}(l)$. As such, we minimize both by minimizing $l$. \replacementpolicy imposes the constraint that the number of logical partitions in the buffer $c_l \ge 2$ and that $p/c = l/c_l$. Therefore $l = 2 * p/c$.

\section{Evaluation}
\label{sec:eval}
We implemented \systemname in 16k lines of C++ and 5k lines of Python. We evaluate \systemname on four large-scale graphs (see Table~\ref{tab:intro_table} for dataset statistics), including two from the OGB large-scale challenge~\cite{hu2021ogblsc}, and compare against the popular SoTA GNN systems DGL and PyG. We show that:
\begin{enumerate}[leftmargin=*]
    \item \systemname reaches the same level of accuracy 2-8$\times$ faster and 8-64$\times$ cheaper than DGL and PyG on all datasets for both node classification and link prediction. 
    \item \datastructure allows \systemname to reduce mini-batch sampling and compute times by up to 14$\times$ and 8$\times$ respectively.
    \item \systemname enables cheap and efficient training of GNNs using disk storage. \replacementpolicy yields runtime and accuracy improvements compared to SoTA methods. Additionally, \systemname allows us to train GNNs for node classification when datasets exceed commodity main memory. 
    \item \systemname auto-tuning rules for \replacementpolicy yield configurations that achieve the highest throughput and model quality, lowering the deployment burden for training. 
\end{enumerate}

\subsection{Experimental Setup}
\label{sec:exp_setup}
We discuss the setup used throughout the experiments.

\newparagraph{Baselines}
We compare end-to-end GNN training over large-scale graphs in \systemname against DGL 0.7 and PyG 2.0.3 (late 2021 releases) (Section~\ref{subsec:e2e_comp}). In addition to end-to-end performance, we evaluate the effect of \datastructure on training by measuring the time for multi-hop sampling and GNN forward/backward pass computation in these systems and \systemname (Section~\ref{subsec:dense_eval}). Furthermore, we compare the multi-hop sampling time in \systemname to the SoTA sampling implementation in NextDoor~\cite{nextdoor} (Section~\ref{subsec:dense_eval}). We do not use NextDoor for end-to-end GNN training as their open-source release supports only limited GNN computation over small graphs which fit in GPU memory.

While \systemname uses techniques such as a partition buffer and CPU-GPU pipelining, which are employed in prior disk-based graph systems (e.g., Marius~\cite{mohoney2021marius} and PyTorch BigGraph~\cite{pytorchbiggraph}), we do not compare against these systems directly as they do not support GNN computation. The primary reason for this is that neither system contains the neighborhood sampling algorithms and data structures needed to support multi-layer GNNs. As described in this work, \systemname addresses the limitation of these systems by developing an efficient neighborhood sampler (\datastructure). Moreover, while both \systemname and Marius use a partition buffer for disk-based training (Marius' disk-based training approach was shown to outperform that of PyTorch BigGraph), each system uses a different partition replacement policy. In Section~\ref{sec:eval_disk_pol} we compare the \replacementpolicy policy developed in \systemname for high throughput and high accuracy disk-based GNN training with Marius' partition replacement policy (by implementing the later in \systemname).

\newparagraph{Hardware Setup}
We evaluate all systems using AWS P3 instances (Table \ref{tab:experiment_setup}). We use an EBS volume with 1GBps of bandwidth and 10000 IOPS as disk storage. We report results for \systemname using two hardware configurations: one for disk-based training (M-GNN$_{Disk}$) and one for training with the full graph in memory (M-GNN$_{Mem}$). For the former, we minimize training costs by using the P3.2xLarge machine---an instance that does not have enough CPU memory to store any of the large-scale graphs in Table~\ref{tab:intro_table}. For the latter, we use the cheapest P3 instance which has enough RAM for training (either a P3.8xLarge or P3.16xLarge). Baseline systems do not support training if graph data does not fit in CPU memory. Thus, for each graph, we report results for DGL and PyG using the same P3 instance that was used for M-GNN$_{Mem}$. We allow baseline systems to use the maximum number of GPUs they support and available in the instance. \systemname uses only one GPU for all experiments.

\begin{table}[t]\footnotesize
    \caption{Cloud GPU instances used for experiments.}
    \vspace{-0.1in}
    \label{tab:experiment_setup}
    \begin{tabular}{l c c c c c}
        \toprule
        AWS Machine & (\$/hr) & GPUs & CPUs & CPU Mem (GB) \\
        \midrule
        P3.2xLarge & 3.06 & 1 & 8 & 61 \\
        P3.8xLarge & 12.24 & 4 & 32 & 244 \\
        P3.16xLarge & 24.48 & 8 & 64 & 488 \\
        \bottomrule
    \end{tabular}
    \vspace{-0.1in}
\end{table}

\newparagraph{Node Classification: Datasets, Models, and Metrics}
We use the two largest OGB node classification graphs: Mag240M and Papers100M (Papers)~\cite{hu2020ogb, hu2021ogblsc}. For Mag240M we use only the paper nodes and citation edges, denoted as Mag240M-Cites (Mag). Based on the graph memory overheads, Papers100M and Mag240M-Cites require a P3.8xLarge and P3.16xLarge respectively for in-memory training. We train a three-layer GraphSage (GS)~\cite{graphsage} GNN on both datasets, a common choice for these graphs~\cite{kaler2022accelerating, zheng2022distributed}. We use 30, 20, and 10 neighbors per layer (ordered away from the target nodes) and sample from both incoming and outgoing edges. We report multi-class classification accuracy averaged over three runs and train for ten epochs. We find that PyG multi-GPU training runs out of CPU memory for Mag240M-Cites, hence, for PyG on this dataset we switch to single-GPU training. 

\newparagraph{Link Prediction: Datasets, Models, and Metrics}
For link prediction, we use the largest OGB link prediction graph---WikiKG90Mv2 (Wiki)~\cite{hu2021ogblsc}. As a second graph for large-scale link prediction, we use Freebase86M (FB)~\cite{zheng2020dglke}. Both datasets fit in CPU memory on an AWS P3.8xLarge machine. We train a GraphSage GNN on both datasets, and the more computationally expensive GAT~\cite{gat} on Freebase86M (the smaller dataset). Both GNNs use a single layer. We use 20 neighbors sampled from incoming and outgoing edges for GraphSage and 10 incoming neighbors for GAT. We evaluate the accuracy of link prediction models using the commonly reported MRR metric~\cite{mohoney2021marius, zheng2020dglke, pytorchbiggraph} using the DistMult~\cite{distmult} score function and train all systems for a fixed number of epochs: five on Freebase86M and ten on WikiKG90Mv2. We report the MRR for a single run due to cost considerations, but report runtime averaged across all training epochs.

Both DGL and PyG provide limited support for training link prediction at scale: PyG does not provide a negative sampler. We implemented negative sampling in PyG based on the negative sampling used in \systemname. DGL provides a negative sampler but the implementation limits the amount of negative samples that can be used to train in a reasonable amount of time. As such, for DGL we use five times fewer negative samples per training edge compared to \systemname to prevent GPU out-of-memory issues. We find that neither baseline supports multi-GPU training for this task: the data loader implementation for link prediction in PyG supports only a single GPU and DGL's multi-GPU training ran out of CPU memory on the AWS P3.8xLarge.

\newparagraph{Hyperparameters}
We use the same values for hyperparameters which define the GNN model and training process across systems. We choose these values to be those used by OGB or prior works~\cite{hu2020ogb, hu2021ogblsc, graphsage} to achieve high accuracy on each dataset. However, to prevent GPU out-of-memory, for PyG on Mag240M-Cites, we use a smaller batch size (half) than DGL and \systemname. While we make sure to request the same number of neighbors per layer for each system, differences in mini batches are expected due to the use of different sampling algorithms. For throughput parameters specific to each system and independent of the computation (e.g., the number of data loader threads), we tune each system and use the best configuration. \systemname disk-based training hyperparameters are set using the auto-tuning rules (see Section~\ref{sec:hyperparameters}). The specific hyperparameters used for each experiment can be found in our artifact (see the Appendix).


\subsection{End-to-End System Comparisons}
\label{subsec:e2e_comp}
We discuss end-to-end training results for \systemname, DGL, and PyG on node classification and link prediction tasks.

\begin{figure}
  \centering
  \includegraphics[width=.41\textwidth]{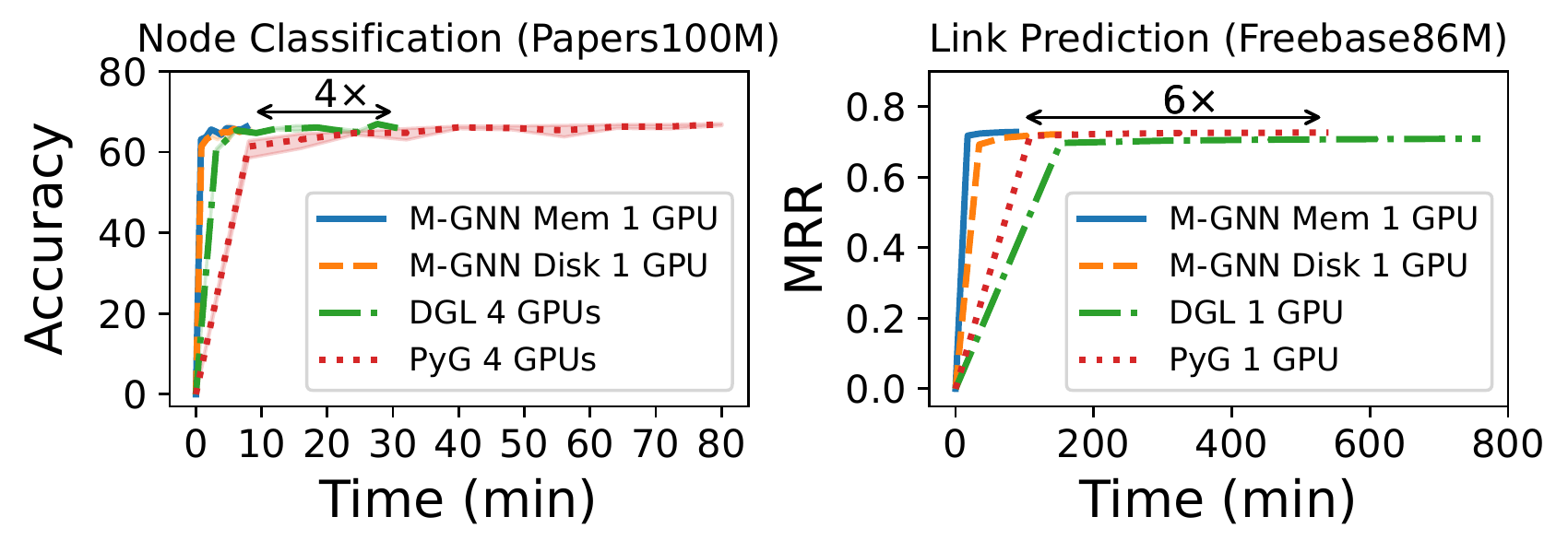}
  \vspace{-0.15in}
  \caption{Time-to-accuracy for \systemname, DGL, and PyG. \systemname reaches the same level of accuracy 4-6$\times$ faster.}
  \label{fig:convergence_papers}
  \vspace{-0.1in}
\end{figure}

Results are reported in Tables~\ref{tab:sys_comparison_nc}-\ref{tab:sys_comparison_fb_model} and Figure~\ref{fig:convergence_papers}. For each experiment we train all systems for the same fixed number of epochs and measure 1) the per-epoch runtime, 2) model accuracy or MRR, and 3) the monetary cost per epoch based on AWS pricing. We report two configurations for \systemname---one with graph data stored in main memory ($\text{M-GNN}_{Mem}$) and one using disk-based training ($\text{M-GNN}_{Disk}$). Next, we highlight key takeaways among all end-to-end results before focusing on each setting (Table) in more detail. 

\newparagraph{Key Takeaway}
\systemname provides the fastest \textit{and} cheapest training option to comparable accuracy for all dataset and model combinations on both learning tasks. Moreover, cost reductions for all experiments are \textit{at least} $8\times$. Differences in training time and cost can be almost two orders of magnitude: Baseline systems can take six days and \$1720 dollars for training (see training on Wiki in Table~\ref{tab:sys_comparison_lp}) yet \systemname needs only eight hours or \$36 dollars for the same dataset. 

\newparagraph{Node Classification}
We focus on end-to-end results for node classification in more detail (Table~\ref{tab:sys_comparison_nc}). With graph data stored in main memory, \systemname with one GPU trains 4$\times$ and 3$\times$ faster than DGL (the fastest baseline) using four and eight GPUs on Papers100M and Mag240M-Cites respectively. All three systems reach similar accuracy on both datasets. We show the time-to-accuracy on Papers100M in Figure~\ref{fig:convergence_papers}. Not only does \systemname train 4$\times$ faster than baseline systems per epoch, but it also lowers the time-to-accuracy by the same factor. There are two reasons for the reduced runtime of \systemname in this setting. First, the \datastructure data structure allows for faster CPU-based mini-batch sampling and GPU-based GNN computation in \systemname compared to baseline systems (evaluated in Section~\ref{subsec:dense_eval}). Second, while both DGL and PyG require multi-GPU machines on AWS due to CPU memory requirements, they both underutilize the additional compute resources: DGL and PyG four-GPU training on Papers100M are only 1.4$\times$ and 1.1$\times$ faster than their single-GPU performance respectively, and DGL eight-GPU training on Mag240M-Cites is only 2.2$\times$ faster than with one-GPU (single-GPU baselines not reported in Table~\ref{tab:sys_comparison_nc}).

While all systems reach comparable accuracy for training with the full graph in memory (within 1\%), \systemname accuracy is 0.55\% and 0.3\% lower than the closest baseline (PyG) on Papers100M and Mag240M-Cites respectively. This is because \datastructure reuses previously sampled one-hop neighbors across layers when constructing multi-layer GNN dataflow graphs. Sample reuse leads to fewer opportunities for one-hop neighborhood randomness resulting in fewer unique nodes in the sampled multi-hop neighborhood for each mini batch (quantified in Table~\ref{tab:mini_batch_benchmarks}). Although this leads to an accuracy reduction for multi-layer GNNs in \systemname, sample reuse is isolated to a single mini batch. Over the course of training, the one-hop neighbors of each node are still randomized across batches, allowing \systemname to achieve comparable accuracy to baselines while training with \datastructure.

\systemname can train the same GraphSage models for node classification using disk-based training on a single AWS P3.2xLarge machine, leading to 16$\times$ and 64$\times$ cheaper learning on the two graphs. Moreover, disk-based node classification in \systemname can actually be faster than in-memory training (e.g., Mag240M-Cites). This occurs because neighborhood sampling operations are performed over in-memory subgraphs, leading to fewer returned neighbors and smaller mini batches. While this can improve throughput, it can also introduce slight accuracy reductions (e.g., 63.17 to 62.53).

\begin{table}[t]\footnotesize
    \caption{\systemname, DGL, and PyG for node classification on large-scale graphs using a GraphSage GNN. Using a single GPU, \systemname can reach the same level of accuracy as multi-GPU baselines 3-8$\times$ faster and up to 64$\times$ cheaper.}
    \vspace{-0.1in}
    \label{tab:sys_comparison_nc}
    \begin{tabular}{l c c c c c c}
        \toprule
        & \multicolumn{2}{c}{Epoch (min.)} &
        \multicolumn{2}{c}{Accuracy} & 
        \multicolumn{2}{c}{Cost (\$/epoch)} \\
        \cmidrule(lr){2-3}
        \cmidrule(lr){4-5}
        \cmidrule(lr){6-7}
        Dataset & Papers & Mag & Papers & Mag & Papers & Mag\\
        \cmidrule(lr){1-7}
        $\text{M-GNN}_{Mem}$ & \textbf{0.77} & 2.57& 66.38 & 63.17 & 0.16 & 1.05\\
        $\text{M-GNN}_{Disk}$ & 0.83 & \textbf{0.94} & 66.03 & 62.53 & \textbf{0.04} & \textbf{0.05}\\
        DGL & 3.07 & 7.83 & 66.98 & 63.73 & 0.63 & 3.19\\
        PyG & 8.01 & 19 & 66.93 & 63.47 & 1.63 & 7.75 \\
        \bottomrule
    \end{tabular}
    \vspace{-0.05in}
\end{table}

\begin{table}[t]\footnotesize
    \caption{\systemname, DGL, and PyG for link prediction on large-scale graphs. All systems use a GraphSage GNN and one GPU. \systemname reaches comparable accuracy to baselines 6$\times$ faster and 13-18$\times$ cheaper. (OOT: out of time)}
    \vspace{-0.1in}
    \label{tab:sys_comparison_lp}
    \begin{tabular}{l c c c c c c}
        \toprule
        & \multicolumn{2}{c}{Epoch (min.)} & 
        \multicolumn{2}{c}{MRR} & 
        \multicolumn{2}{c}{Cost (\$/epoch)} \\
        \cmidrule(lr){2-3}
        \cmidrule(lr){4-5}
        \cmidrule(lr){6-7}
        Dataset & FB & Wiki & FB & Wiki & FB & Wiki\\
        \cmidrule(lr){1-7}
        $\text{M-GNN}_{Mem}$ & \textbf{17.5} & \textbf{46.6} & .7285 & .4655 & 3.57 & 9.38\\
        $\text{M-GNN}_{Disk}$ & 34.2 & 69.9 & .7216 & .4156 & \textbf{1.74} & \textbf{3.56}\\
        DGL & 152 & 844 & .7091 & OOT & 31.0 & 172\\
        PyG & 108 & 312 & .7267 & .4683 & 22.0 & 63.6 \\
        \bottomrule
    \end{tabular}
    \vspace{-0.05in}
\end{table}

\begin{table}[t]\footnotesize
    \caption{Comparison of GraphSage (GS) and GAT GNN training in \systemname, DGL, and PyG for link prediction on Freebase86M. Baselines bottlenecked by CPU-based mini batch construction result in similar training time and cost on GraphSage and the more computationally expensive GAT.}
    \vspace{-0.1in}
    \label{tab:sys_comparison_fb_model}
    \begin{tabular}{l c c c c c c}
        \toprule
        & \multicolumn{2}{c}{Epoch (min.)} & 
        \multicolumn{2}{c}{MRR} & 
        \multicolumn{2}{c}{Cost (\$/epoch)} \\
        \cmidrule(lr){2-3}
        \cmidrule(lr){4-5}
        \cmidrule(lr){6-7}
        Model & GS & GAT & GS & GAT & GS & GAT\\
        \cmidrule(lr){1-7}
        $\text{M-GNN}_{Mem}$ & \textbf{17.5} & \textbf{52.6} & .7285 & .7331 & 3.57 & 10.7\\
        $\text{M-GNN}_{Disk}$ & 34.2 & 56.9 & .7216 & .7251 & \textbf{1.74} & \textbf{2.90}\\
        DGL & 152 & 151 & .7091 & .6516 & 31.0 & 30.8\\
        PyG & 108 & 107 & .7267 & .7252 & 22.0 & 21.8\\
        \bottomrule
    \end{tabular}
    \vspace{-0.05in}
\end{table}

\begin{table*}[t]\footnotesize
    \caption{Comparison of the time required for mini-batch neighborhood sampling, GPU-based computation, and the number of nodes/edges sampled per mini batch in \systemname, DGL, and PyG for GraphSage GNNs of varying depth on Papers100M. Using \datastructure, in \systemname sampling is 14$\times$ and GPU computation is 8$\times$ faster for a four-layer GNN. These speedups occur in part because \datastructure allows \systemname to sample fewer nodes/edges to construct mini batches. (OOM: out of memory)}
    \vspace{-0.1in}
    \label{tab:mini_batch_benchmarks}
    {
        \begin{tabular}{l c c c c c}
            \toprule
            & \multicolumn{5}{c}{CPU Sampling Time (ms)} \\
            \cmidrule(lr){2-6} 
            \#Layers & 1 & 2 & 3 & 4 & 5 \\
            \cmidrule(lr){1-6} 
            M-GNN & \textbf{1.4} & \textbf{18} & \textbf{103} & \textbf{401} & \textbf{1.8k}\\
            DGL & 5.7 & 28 & 376 & 5.4k & 49k\\
            PyG & 2.2 & 59 & 1227 & 19k & 96k\\
            \bottomrule
        \end{tabular}
    }
    \hfill
    {
        \begin{tabular}{l c c c c c}
            \toprule
            \multicolumn{5}{c}{GPU Computation Time (ms)} \\
            \cmidrule(lr){1-5} 
            1 & 2 & 3 & 4 & 5 \\
            \cmidrule(lr){1-5} 
            4 & \textbf{6.1} & \textbf{21} & \textbf{153} & OOM\\
            4.7 & 29 & 215 & 1231 & OOM \\
            \textbf{3.2} & 13 & 168 & OOM & OOM \\
            \bottomrule
        \end{tabular}
    }
    \hfill
    {
        \begin{tabular}{p{0.26in} c c c c c}
            \toprule
            \multicolumn{5}{c}{Number of Nodes/Edges Sampled Per Mini Batch} \\
            \cmidrule(lr){1-5} 
            1 & 2 & 3 & 4 & 5 \\
            \cmidrule(lr){1-5} 
            12k/13k & 136k/181k & 1M/2M & 6M/17M & 23M/91M\\
            13k/20k & 182k/278k & 2M/4M & 9M/37M & 33M/222M\\
            13k/20k & 178k/258k & 2M/4M & 9M/32M & 31M/174M\\
            \bottomrule
        \end{tabular}
    }
\end{table*}

\newparagraph{Link Prediction}
We now focus on link prediction. End-to-end results for all systems on Freebase86M and WikiKG90Mv2 are reported in Table~\ref{tab:sys_comparison_lp}. Time-to-accuracy on Freebase86M is shown in Figure~\ref{fig:convergence_papers}. \systemname in-memory training is 6$\times$ and 7$\times$ faster than the best baseline on the two datasets respectively. While PyG and \systemname reach comparable model quality, DGL is lower due to its use of fewer negative samples. On WikiKG90Mv2, DGL does not complete the ten training epochs within two days. To compare system performance for different models, we report results for GraphSage and GAT GNNs on Freebase86M in Table~\ref{tab:sys_comparison_fb_model}. Interestingly, DGL and PyG exhibit similar runtimes for GraphSage and the more computationally expensive GAT. This result supports the fact that baseline systems are bottlenecked by CPU sampling operations rather than GPU computation. 

\replacementpolicy allows \systemname to train the same models for link prediction on a 3$\times$ cheaper P3.2xLarge machine by utilizing disk storage. For this task, disk-based training in \systemname is slower than in-memory training for two reasons: 1) \replacementpolicy requires performing disk IO during every epoch, and 2) the P3.2xLarge machine has 4$\times$ fewer CPU resources available for reading/writing embeddings to main memory. Yet, epoch runtimes remain $1.9\times$-$4.5\times$ faster than baseline systems, yielding cost reductions of 7.5-18$\times$. As described in the introduction, achieving high-accuracy disk-based GNN models for link prediction is a key challenge. On Freebase86M, \replacementpolicy allows \systemname to reach comparable model quality to the in-memory setting. Yet, recovering in-memory accuracy remains an open problem for some datasets (Wiki). We evaluate \replacementpolicy in more detail in Section~\ref{sec:eval_disk_pol} and show that it trains faster while improving accuracy compared to SoTA policies for seven model/dataset combinations.

\subsection{Extreme Scale GNN Training With One GPU}
\label{subsec:extreme_scale}
A core motivation of our work is to investigate if distributed GNN training~\cite{zheng2022distributed, distDGL} is necessary or if the resources in a single machine can be used efficiently to scale GNN training. To evaluate this, we stress-test \systemname with respect to graph size: We consider the task of learning vector representations for link prediction over the \textit{entire hyperlink graph from the Common Crawl 2012 web corpus}, a graph with 3.5 billion nodes (web pages) and 128 billion edges (hyperlinks between pages) (Table~\ref{tab:intro_table}). We use \systemname disk-based training on an AWS P3.2xLarge instance with one GPU, 60GB of RAM, and 4TB of SSD storage. To learn the representations, we use a GraphSage GNN with 10 neighbors, the DistMult score function with 500 negative samples, and an embedding dimension of 50. We find that \systemname is able to train this GNN model over the hyperlink graph---a graph with 210$\times$ more edges than the largest graph in the OGB large-scale challenge~\cite{hu2021ogblsc} (WikiKG90Mv2)---while maintaining a throughput of 194k edges/sec, leading to a monetary cost of only \$564 per epoch. Thus, \systemname costs only 3.3$\times$ more per epoch while training on the hyperlink graph compared to baseline systems training on WikiKG90Mv2. This experiment presents an initial large-scale benchmark that can be used by the community to measure the cost of distributed training over large clusters and to understand the power of optimized single machine deployments.

\subsection{Effect of \datastructure on Training}
\label{subsec:dense_eval}
We have shown that end-to-end training in \systemname is faster than existing systems. The key reason for this result is the efficient mini-batch sampling and forward pass computation using the \datastructure data structure. In this section, we report the effect of \datastructure on training. Recall that training consists of two phases: 1) CPU-based mini batch construction via neighborhood sampling and 2) GPU-based GNN forward/backward pass computation. While \datastructure is co-designed for both efficient sampling and GNN computation, we seek to understand the effect of \datastructure on each individual training phase. As a result, we measure the average time per mini batch for 1) CPU neighborhood sampling and 2) GPU training and compare those times against the corresponding methods used in DGL and PyG. For these experiments, we use a GraphSage GNN on the Papers100M dataset and vary the number of GNN layers from one to five. For each layer, we request a max of 10 incoming and 10 outgoing neighbors per node from each system. We use the same hyperparameters for \systemname, DGL, and PyG and train all systems with the graph in main memory using one GPU. 

We report the average CPU-based neighborhood sampling time for each system in Table~\ref{tab:mini_batch_benchmarks}. \datastructure allows \systemname to sample multi-hop neighborhoods faster than baseline systems for all configurations. For three, four, and five layers, \systemname is 3.7$\times$, 14$\times$, and 26$\times$ faster than the best baseline. GNN training on the GPU in \systemname is also faster than DGL and PyG (Table~\ref{tab:mini_batch_benchmarks}). \datastructure leads to 8$\times$ faster computation compared to the best baseline for three- and four-layer GNNs. We find that for five-layer GNNs, mini batches become too large and cause all three systems to run out of memory on the AWS NVIDIA V100 GPUs with 16GB of memory (but could be used on new GPUs with 80GB). 

\begin{table}[t]\footnotesize
    \caption{Comparison of the time required for GPU-based mini-batch neighborhood sampling in \systemname and NextDoor for GraphSage GNNs of varying depth on LiveJournal. Sample reuse in \datastructure leads to better scaling with respect to the number of GNN layers, allowing \systemname to outperform optimized sampling implementations.}
    \vspace{-0.1in}
    \label{tab:nextdoor}
    \begin{tabular}{l c c c c c}
        \toprule
        & \multicolumn{5}{c}{GPU Sampling Time (ms)} \\
        \cmidrule(lr){2-6} 
        \#Layers & 1 & 2 & 3 & 4 & 5 \\
        \cmidrule(lr){1-6} 
        M-GNN & 1 & 2.5 & 9.6 & 25 & 32\\
        NextDoor & 0.1 & 0.5 & 6.5 & 135 & OOM\\
        \bottomrule
    \end{tabular}
    \vspace{-0.1in}
\end{table}

We investigate to what extent the sampling and computation improvements in \systemname can be attributed to the reuse of neighborhood samples in \datastructure compared to implementation co-design choices, i.e., parallel sampling on the CPU and dense kernels on the GPU. In Table~\ref{tab:mini_batch_benchmarks}, we report the average number of unique nodes and edges sampled per mini batch for each system. \datastructure allows for mini batch construction using fewer samples than baselines. For example, constructing a three-hop neighborhood in \systemname requires sampling half as many nodes and edges compared to DGL and PyG (for the same number of target nodes). While sample reuse in \datastructure is evident, mini batch sizes in \systemname, DGL, and PyG are all the same order of magnitude, yet \systemname sampling and computation improvements are more significant (e.g., 14$\times$ and 8$\times$). This result validates the co-design of \datastructure: parallel CPU sampling algorithms and the use of dense GPU kernels, together with one-hop sample reuse, lead to the improved throughput in \systemname.

\newparagraph{Comparison Against Accelerated Sampling Kernels}
To further evaluate the benefit of \datastructure on GNN training, we compare the multi-hop sampling in \systemname to the SoTA accelerated sampling implementation of NextDoor~\cite{nextdoor}. NextDoor uses GPUs to reduce sampling times and employs optimized GPU kernels for parallelization, load balancing, and caching. These kernels allow NextDoor to outperform multi-hop sampling implementations in existing systems, but their open-source release requires that graphs fit in GPU memory.
While in \systemname we focus primarily on CPU-based multi-hop sampling for mixed CPU-GPU training to scale to large graphs (as discussed throughout the paper), \systemname also includes support for GPU-based multi-hop sampling for end-to-end training on smaller graphs without CPU involvement. Unlike our CPU-based sampling implementation which uses optimized parallel algorithms to construct \datastructure, our GPU-based sampling implementation builds \datastructure using only default PyTorch functions. We compare GPU-based sampling using \datastructure in \systemname to the optimized sampling kernels in NextDoor by measuring the average multi-hop sampling time per mini batch for a GraphSage GNN of varying depth on the LiveJournal dataset (which fits in GPU memory with 4.8M nodes/69M edges)~\cite{livejournal}. For each layer, we sample 20 outgoing neighbors per system.

GPU-based sampling times for \systemname and NextDoor are shown in Table~\ref{tab:nextdoor}. The optimized sampling kernels in NextDoor have lower overhead and better parallelization for one-hop sampling compared to the default PyTorch functions used by \systemname. These kernels lead to faster sampling for one- and two-layer GNNs. For deeper GNNs however, \systemname is comparable to or faster than NextDoor. This is because as the number GNN layers increases \datastructure has more opportunities to minimize redundant one-hop sampling compared to NextDoor by reusing previous samples across layers. Table~\ref{tab:nextdoor} shows that redundant sampling can bottleneck even the most optimized sampling implementations. At the same time, \datastructure avoids this bottleneck and can scale to five-layer GNNs with little sampling overhead.

For graphs that exceed GPU memory, the open-source release of NextDoor is unable to perform multi-hop sampling. In this setting, \systemname uses mixed CPU-GPU training with CPU-based neighborhood sampling (evaluated above).

\begin{table}[t]\footnotesize
    \caption{\replacementpolicy versus the SoTA BETA policy~\cite{mohoney2021marius} for disk-based link prediction using GraphSage (GS) and GAT GNNs as well as the DistMult (DM) knowledge graph embedding model. \replacementpolicy leads to simultaneously faster training and higher MRR. (237: FB15k-237; Epoch time in seconds for 237)}
    \vspace{-0.1in}
    \label{tab:beta_battles}
    \begin{tabular}{l l p{2em} c c c c}
        \toprule
        \multirow{2}{*}[-.5ex]{Model} &
        \multirow{2}{*}[-.5ex]{Graph} &
        \multirow{2}{2em}[-.5ex]{Mem MRR} &
        \multicolumn{2}{c}{Disk-Based MRR} & 
        \multicolumn{2}{c}{Epoch (min.)} \\
        \cmidrule(lr){4-5} 
        \cmidrule(lr){6-7}
        &&& \replacementpolicy & BETA & \replacementpolicy & BETA\\
        \cmidrule(lr){1-7}
        DM & 237 & .2533 & \textbf{.2659} & .2431 & \textbf{1.78} & 1.95\\
        DM & FB & .7249 & \textbf{.7220} & .7189 & \textbf{13.73} & 17.51\\
        DM & Wiki & .3941 & \textbf{.4071} & .3951 & \textbf{22.54} & 27.75\\
        \cmidrule(lr){1-7}
        GS & 237 & .2825 & \textbf{.2736} & .2369& \textbf{3.07} & 3.28\\
        GS & FB & .7342 & \textbf{.7123} & .6976 & \textbf{47.45} & 50.08\\
        GS & Wiki & .4658 & .4078 & .4080 & \textbf{76.66} & 82.34\\
        \cmidrule(lr){1-7}
        GAT & 237 & .2869 & \textbf{.2341} & .2076 & \textbf{3.51} & 3.90\\
        GAT & FB & .7418 & \textbf{.7053} & .6860 & \textbf{42.01} & 46.02\\
        \bottomrule
    \end{tabular}
    \vspace{-0.1in}
\end{table}

\subsection{Evaluating \replacementpolicy for Disk-Based Training}
\label{sec:eval_disk_pol}
In Section~\ref{subsec:e2e_comp}, we showed that \replacementpolicy allows for disk-based GNN training on the link prediction task 7.5-18$\times$ cheaper than DGL and PyG. We now evaluate \replacementpolicy in more detail and compare to the SoTA greedy policy from Marius~\cite{mohoney2021marius} called BETA. We perform disk-based training using both methods and measure 1) the per-epoch runtime and 2) the disk-based model accuracy (using MRR). We also report MRR for in-memory training as a baseline for disk-based MRR. Since Marius does not support GNNs, we implement BETA in \systemname. We use the GraphSage and GAT GNNs on the graphs FB15k-237~\cite{fb15k237}, Freebase86M, and WikiKG90Mv2. We include FB15k-237 (14541 nodes, 272115 edges) to measure the bias present in disk-based training policies while utilizing all neighbors for GNN aggregation and all negatives for computing MRR (as opposed to using neighbor/negative sampling for large graphs). We also use the DistMult knowledge graph embedding model to compare \replacementpolicy and BETA on the specialized decoder-only models supported by Marius. We utilize a buffer capacity that can store 1/4 of all partitions in memory. We enable prefetching to overlap IO with computation. \replacementpolicy hyperparameters are set as described in Section~\ref{sec:hyperparameters}. For BETA, which has no auto-tuning rules, we manually tune the number of partitions for best performance.

We report the runtime and MRR for all models and datasets in Table~\ref{tab:beta_battles}. While the BETA policy achieves near in-memory MRR for the specialized DistMult model, MRR drops by up to 16\% for GraphSage and GAT GNNs. By promoting mini-batch randomness, \replacementpolicy reduces the gap to in-memory training for GNN models by up to 80\%. Moreover, \replacementpolicy actually results in improved disk-based MRR for DistMult as well. Overall, \replacementpolicy results in higher MRR compared to BETA for seven of the eight model/dataset combinations. Yet completely recovering the in-memory MRR for disk-based link prediction remains a challenge (e.g., GAT on FB15k-237 or GS on Wiki) and area of interest for future work.

While \replacementpolicy allows for higher MRR compared to BETA, it also simultaneously allows for faster training. In particular, epoch time is reduced for DistMult---a less compute intensive model---which is IO bottlenecked. For example, \replacementpolicy is 1.28$\times$ faster than BETA for DistMult on Freebase86M. In this setting, prefetching to overlapping IO with computation is needed for high throughput. While both \replacementpolicy and BETA minimize IO, by decoupling mini batch generation from partition replacement and allowing for the deferred processing of training examples, \replacementpolicy evenly distributes mini batches and IO across each epoch. This is in contrast to the greedy BETA policy which results in most mini batch processing occurring during the early part of each epoch, leaving little computation to hide IO for the latter part.

\subsection{Evaluating \replacementpolicy Auto-tuning Rules}
\begin{figure}
  \centering
  \includegraphics[width=.46\textwidth]{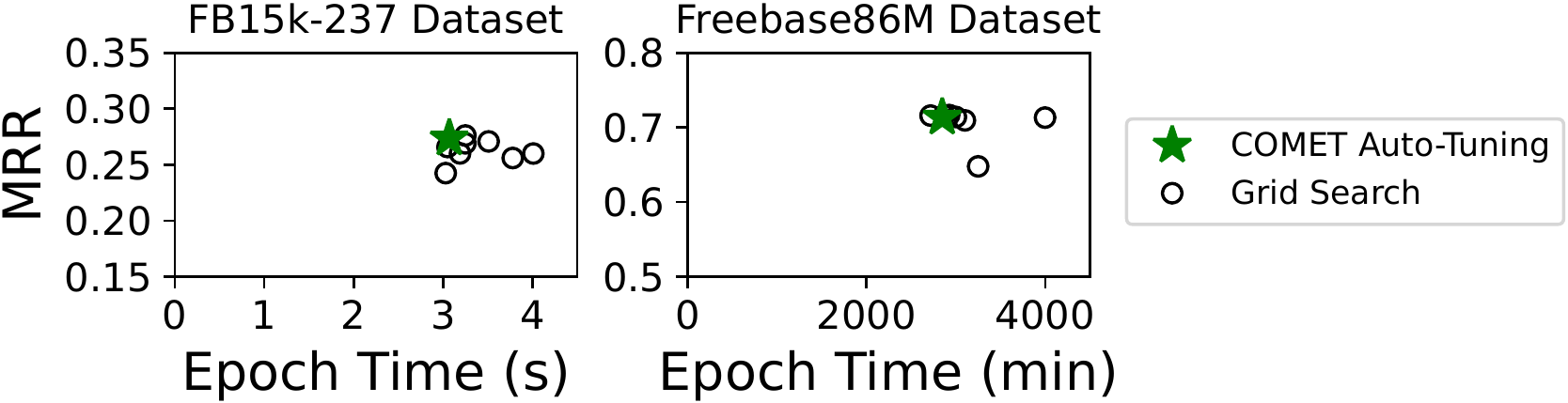}
  \vspace{-0.1in}
  \caption{MRR and runtime for GraphSage GNN disk-based training with \replacementpolicy using different hyperparameters. The auto-tuning rules used by \systemname are near-optimal.}
  \label{fig:beta_hyperparameter_scan}
  \vspace{-0.1in}
\end{figure}

We evaluate the effectiveness of the parameter auto-tuning rules used in \systemname for disk-based training. To this end, we measure the runtime and MRR of \replacementpolicy obtained when training uses the rules described in Section~\ref{sec:hyperparameters} and compare against the runtime and MRR obtained for each configuration in a hyperparameter scan. We use a GraphSage GNN and train on two datasets (FB15k-237 and Freebase86M). Results are shown in Figure~\ref{fig:beta_hyperparameter_scan}. The auto-tuning rules used by \systemname lead to a hyperparameter setting that achieves near-optimal runtime and MRR simultaneously, eliminating the need for expensive hyperparameter search. 

\section{Related Work}

\newparagraph{Systems for ML over Graph Data} 
Many systems support GPU training of GNNs~\cite{p3gnn, MLSYS2020_ROC, kaler2022accelerating, aligraph, pagraph, dong2021global, gnnsurvey}. Two such popular systems are DGL~\cite{dgl} and PyTorch Geometric~\cite{pyg}. Complementary to these systems, many works focus on scaling different dimensions of GNN training: To reduce the overhead of mixed CPU-GPU training, some works highlight the importance of GPU-oriented data communication or caching~\cite{largegcn, min2021pytorchdirect, kaler2022accelerating, pagraph, dong2021global}. Additional works focus on optimized GPU kernels~\cite{gnnadvisor}. In general, these works focus on orthogonal challenges of GNN training than those discussed here and these ideas can be incorporated into \systemname. Finally, there are works that focus on scaling the training of non-GNN based graph ML models~\cite{mohoney2021marius,  zheng2020dglke, pytorchbiggraph, gosh}. For example, Marius~\cite{mohoney2021marius} utilizes pipelined training to achieve state-of-the-art throughput for specialized knowledge graph embeddings.

\newparagraph{Large-Scale Training}
To scale GNN training to graphs that exceed the CPU memory capacity of a single machine, many works opt for a distributed multi-machine approach~\cite{zheng2022distributed, p3gnn, MLSYS2020_ROC, aligraph, wang2021flexgraph}. Recent work introduces DistDGLv2 as a distributed version of DGL~\cite{zheng2022distributed} and utilizes METIS partitioning, co-location of data with mini batch computation, and asynchronous mini batch preparation to scale training. Other works distribute training in a serverless manner~\cite{dorylus}. In \systemname, we use a disk-based approach to scaling beyond CPU memory. On Papers100M, \systemname is 6.7$\times$ cheaper than DistDGLv2 based on the reported cost to 66\% accuracy for each system. Other works have previously supported disk-based training for link prediction using non-GNN models~\cite{partitioning, mohoney2021marius, pytorchbiggraph}. \systemname provides disk-based GNN support for both node classification and link prediction. 

\newparagraph{Neighborhood Sampling}
Many works focus on reducing the overhead of neighborhood sampling. Initial approaches sample a fixed number of neighbors per node~\cite{graphsage}, while follow-up works sample a fixed number of neighbors per layer~\cite{chen2018fastgcn, zou2019layer}. Other works decouple the sampling frequency from the mini batch frequency~\cite{lazygcn}. \systemname focuses on sampling a fixed number of neighbors per node with minimal redundancy. Still other works focus on making mini-batch training more efficient by increasing the density of edges between nodes in a mini batch~\cite{zeng2020graphsaint, clustergcn}. These contributions can be incorporated in \systemname and are orthogonal to our study. Finally, recent works utilize GPUs to speed up sampling~\cite{dong2021global, nextdoor}. \systemname supports GPU-based sampling but uses CPU-based sampling to scale to large graphs.

\section{Conclusion}
\label{sec:conclusion}
This paper introduced \systemname, a system for pipelined mini-batch training of GNNs in a single machine. We showed that \systemname with one GPU can achieve the same level of model accuracy up to 8$\times$ faster than existing systems using eight GPUs. To achieve these results, we introduced the \datastructure data structure to minimize the redundancy of multi-hop sampling and the two-level \replacementpolicy replacement policy for disk-based training. Overall, our results highlight the need to optimize single GPU implementations of ML systems before resorting to multi-GPU approaches.

\begin{acks}
We would like to thank the anonymous reviewers and our shepherd Rong Chen for their constructive comments that helped improve our paper.
This work was supported by NSF under grant 1815538 and DARPA under grants ASKE HR00111990013 and ASKEM HR001122S0005. The U.S. Government is authorized to reproduce and distribute reprints for Governmental purposes notwithstanding any copyright notation thereon. Any opinions, findings, and conclusions or recommendations expressed in this material are those of the authors and do not necessarily reflect the views, policies, or endorsements, either expressed or implied, of DARPA or the U.S. Government. This work is also supported by the National Science Foundation grant CNS-1838733 and by the Office of the Vice Chancellor for Research and Graduate Education at UW-Madison with funding from the Wisconsin Alumni Research Foundation.
\end{acks}

\balance
\bibliographystyle{ACM-Reference-Format}
\bibliography{sections/ref}

\newpage
\clearpage
\appendix
%
\pagebreak

\appendix
\section{Artifact Appendix} 

\subsection{Abstract}
We have released the artifact for reproducing the paper's experimental results in our GitHub repository artifact branch (\texttt{eurosys\_2023\_artifact}). The artifact includes the \systemname source code, Deep Graph Library and PyTorch Geometric baseline implementations, configuration files, and Python scripts to execute the experiments reported in the paper for all three systems. The source code and experiment configurations can be used to study implementation details that were not mentioned in the paper for brevity. Details on how to use the artifact can be found in the following sections and in the GitHub README file.

\subsection{Description \& Requirements}

\subsubsection{How to access}
The artifact is available at the following GitHub url:
\href{https://github.com/marius-team/marius/tree/eurosys_2023_artifact}{\texttt{\systemname GitHub Artifact}}.

\subsubsection{Hardware dependencies}
Artifact hardware dependencies and the specific hardware used for each experiment reported in the paper are provided in the GitHub repository README file. Paper experiments were run on AWS P3 GPU machines to support training over large-scale graphs, but \systemname and the artifact can be run on CPU only hardware as well.

\subsubsection{Software dependencies} 
Artifact software dependencies and the specific versions used for the paper experiments are listed in the GitHub repository README file under the `Build Information and Environment' heading.

\subsubsection{Benchmarks} 
The artifact uses publicly available GNN benchmark datasets (graphs): FB15k-237, OGBN-Arxiv, OGBN-Papers100M, OGB-Mag240M, Freebase86M, and OGB-WikiKG90Mv2. Scripts to download and preprocess all datasets are included in the artifact and fully automated.

\subsection{Set-up}
Installation and configuration steps required to prepare the artifact environment are described in the `Getting Started' section of the GitHub repository README file. In the section titled `End-to-End Docker Installation' we have provided a Dockerfile and instructions which include all dependencies and commands necessary to install and build the artifact. 

\subsection{Evaluation workflow}
In this section, we provide a brief outline of the artifact functionality and intended use of this artifact to reproduce the paper's main claims. Reproducing all experiments reported in the paper requires significant compute resources: The experiments in the paper were run on AWS P3 GPU machines and we estimate that reproducing all the claims may cost approximately \$5000. \textit{Thus, we have organized our artifact according to 1) a `minimal working example' to demonstrate artifact functionality and 2) a list of experiments for reproducing the numbers reported in the paper and the estimated cost of each experiment}. We first highlight the minimal working example that can establish artifact functionality, then list the paper's major claims and how each are supported by the artifact experiments.

\subsubsection{Artifact Functionality}
To demonstrate how the artifact can be used to produce experimental results comparing \systemname, DGL, and PyG, we include a set of minimal working examples described in the GitHub README file under the section `Artifact Minimal Working Example (Functionality)'. These examples consist of Python scripts to train GNNs using each system for the tasks of link prediction (on the FB15k-237 graph) and node classification (on the OGBN-Arxiv graph). The Python scripts follow the same format as those used to create the paper's experiments, but operate on small graphs (rather than the large-scale graphs used in the paper) and can be run without the need for GPUs.

\subsubsection{Major Claims}
The major experimental claims made in the paper are:

\begin{itemize}
    \item (C1): For the task of node classification, \systemname can train GNNs over large-scale graphs 3-8$\times$ faster and up to 64$\times$ cheaper than competing systems while reaching comparable accuracy. This is proven by the experiment (E1) described in Section~\ref{subsec:e2e_comp} with results reported in Table~\ref{tab:sys_comparison_nc}.
    \item (C2): For the task of link prediction, \systemname can train GNNs over large-scale graphs 6$\times$ faster and 13-18$\times$ cheaper than competing systems while reaching comparable accuracy. This is proven by the experiment (E2) described in Section~\ref{subsec:e2e_comp} with results reported in Table~\ref{tab:sys_comparison_lp}.
    \item (C3): A key reason for the improved end-to-end performance of \systemname is faster neighborhood sampling and GNN forward pass computation. Using the \datastructure data structure, in \systemname sampling is 14$\times$ faster and GPU computation is 8$\times$ faster for a four-layer GNN when compared to baseline systems. This is proven by the experiment (E3) described in Section~\ref{subsec:dense_eval} with results reported in Table~\ref{tab:mini_batch_benchmarks}.
    \item (C4): The \replacementpolicy partition replacement policy for disk-based link prediction leads to simultaneously faster training and higher MRR compared to existing SoTA policies. This is proven by the experiment (E4) described in Section~\ref{sec:eval_disk_pol} with results reported in Table~\ref{tab:beta_battles}.
\end{itemize}

\subsubsection{Experiments}
The artifact contains a table in the GitHub repository README file under `Reproducing Experimental Results' which describes what experiment to run to reproduce the above main claims. Each experiment in the table also contains the required hardware, cost/time estimates, a short explanation, the expected results, and an explicit link to one of the above four claims as well as the corresponding table in the paper. All experiments are run in the same way as the minimal working example and as described in the `Artifact Documentation: Running Experiments' section of the README. That is, the  experiment name is provided to the \texttt{run\_experiment.py} script with any additional desired arguments.




\end{document}